%% file: main.tex
\documentclass[10pt,twocolumn,letterpaper]{article}

\usepackage[pagenumbers]{cvpr} 

\input{preamble}

\definecolor{cvprblue}{rgb}{0.21,0.49,0.74}
\usepackage[pagebackref,breaklinks,colorlinks,citecolor=cvprblue]{hyperref}
\usepackage{graphicx}
\usepackage{amsmath}
\usepackage{amssymb}
\usepackage{booktabs, array}
\usepackage{siunitx}
\usepackage{makecell}
\usepackage{cuted}
\usepackage{gensymb}
\usepackage{multirow}
\usepackage[accsupp]{axessibility}  %
\usepackage[capitalize]{cleveref}
\usepackage{colortbl}
\definecolor{gold}{HTML}{D4AF37}
\definecolor{silver}{HTML}{C0C0C0}
\definecolor{bronze}{HTML}{CD7F32}
\definecolor{first}{HTML}{BC68FF}  
\definecolor{second}{HTML}{8585FF} 
\definecolor{third}{HTML}{9DD0FF}  

\def\paperID{4491} 
\def\confName{CVPR}
\def\confYear{2024}

\title{3DGStream: On-the-Fly Training of 3D Gaussians for Efficient Streaming of Photo-Realistic Free-Viewpoint Videos}

\author{ Jiakai Sun, \quad Han Jiao, \quad Guangyuan Li, \quad Zhanjie Zhang, \quad Lei Zhao\footnotemark[1], \quad Wei Xing\footnotemark[1] \\
Zhejiang University\\
{\tt\small \{csjk, csjh, cslgy, cszzj, cszhl, wxing\}@zju.edu.cn }\\
{\url{https://sjojok.github.io/3dgstream}}
}

\begin{document}
\maketitle
{
\renewcommand{\thefootnote}{\fnsymbol{footnote}}
\footnotetext[1]{Corresponding authors.}
}
\input{3DGStream/0_abstract}   
\input{3DGStream/1_intro}
\input{3DGStream/2_related_work}
\input{3DGStream/3_background}
\input{3DGStream/4_method}
\input{3DGStream/5_evaluation}
\input{3DGStream/6_limitations}
\input{3DGStream/7_conclusion}
\input{3DGStream/X_suppl}

{
    \small
    \bibliographystyle{ieeenat_fullname}
    \bibliography{main}
}

\end{document}

%% file: preamble.tex
\usepackage[dvipsnames]{xcolor}


%% file: 3DGStream/0_abstract.tex
\begin{abstract}
    Constructing photo-realistic Free-Viewpoint Videos (FVVs) of dynamic scenes from multi-view videos remains a challenging endeavor.
    Despite the remarkable advancements achieved by current neural rendering techniques, these methods generally require complete video sequences for offline training and are not capable of real-time rendering.
    To address these constraints, we introduce 3DGStream, a method designed for efficient FVV streaming of real-world dynamic scenes. 
    Our method achieves fast on-the-fly per-frame reconstruction within 12 seconds and real-time rendering at 200 FPS.
    Specifically, we utilize 3D Gaussians (3DGs) to represent the scene.
    Instead of the na\"{i}ve approach of directly optimizing 3DGs per-frame, we employ a compact Neural Transformation Cache (NTC) to model the translations and rotations of 3DGs, markedly reducing the training time and storage required for each FVV frame.
    Furthermore, we propose an adaptive 3DG addition strategy to handle emerging objects in dynamic scenes.
    Experiments demonstrate that 3DGStream achieves competitive performance in terms of rendering speed, image quality, training time, and model storage when compared with state-of-the-art methods.
\end{abstract}
\vspace{-5mm}

%% file: 3DGStream/1_intro.tex
\input{3DGStream/figs_tabs/teaser_1}
\input{3DGStream/figs_tabs/teaser_2}
\section{Introduction}
\label{sec:intro}
Constructing Free-Viewpoint Videos~(FVVs) from videos captured by a set of known-poses cameras from multiple views remains a frontier challenge within the domains of computer vision and graphics.
The potential value and application prospects of this task in the VR/AR/XR domains have attracted much research.
Traditional approaches predominantly fall into two categories: geometry-based methods that explicitly reconstruct
dynamic graphics primitives~\cite{collet2015high,motion2fusion},
and image-based methods that obtain new views through interpolation~\cite{zitnick2004high, broxton2020immersive}.
However, these conventional methods struggle to handle real-world scenes characterized by complex geometries and appearance.

In recent years, Neural Radiance Fields~(NeRFs)~\cite{mildenhall2020nerf}
has garnered significant attention due to its potent capabilities in synthesizing novel views as a 3D volumetric representation.
A succession of NeRF-like works~
\cite{pumarola2021d,
li2022dynerf,
park2023temporal,
park2021hypernerf,
wang2023neuralresidual,
li2022streaming,
park2021nerfies,
TiNeuVox,
kplanes_2023,
li2023dynibar,
li2020neural,
xian2020space} further propelled advancements in constructing FVVs on dynamic scenes.
Nonetheless, the vast majority of NeRF-like FVV construction methods encountered two primary limitations:
(1) they typically necessitate complete video sequences for time-consuming offline training, meaning they can replay dynamic scenes but are unable to stream them,
and (2) they generally fail to achieve real-time rendering, thereby hindering practical applications.

Recently, Kerbl \etal~\cite{kerbl3Dgaussians} have achieved real-time radiance field rendering using 3D Gaussians (3DGs),
thus enabling the instant synthesis of novel views in static scenes with just minutes of training.
Inspired by this breakthrough, we propose 3DGStream, a method that utilizes 3DGs to construct Free-Viewpoint Videos (FVVs) of dynamic scenes.
Specifically, we first train the initial 3DGs on the multi-view frames at timestep 0.
Then, for each timestep $i$, we use the 3DGs of previous timestep $i-1$ as initialization and pass it to a two-stage pipeline.
(1)~In Stage 1, we train a Neural Transformation Cache~(NTC) to model the transformations of 3DGs.
(2)~Then in the Stage 2, we use an adaptive 3DG addition strategy to handle emerging objects by spawning frame-specific additional 3DGs near these objects and optimize them along with periodic splitting and pruning.
After the two-stage pipeline concludes, we use both the 3DGs transformed by the NTC and the additional 3DGs for rendering at the current timestep $i$, with only the former carrying over for initialization of the subsequent timestep.
This design significantly reduces the storage requirements for the FVV, as we only need to store the per-frame NTCs and frame-specific additions, rather than all 3DGs for each frame.

3DGStream is capable of rendering photo-realistic FVVs at megapixel resolution in real-time, boasting exceptionally rapid per-frame training speeds and limited model storage requirements.
As illustrated in \cref{fig:teaser_1,fig:teaser_2}, compared with static reconstruction methods that train from scratch per-frame and dynamic reconstruction methods that necessitate offline training across the complete video sequences,
our approach excels in both training speed and rendering speed, maintaining a competitive edge in image quality and model storage.
Furthermore, our method outperforms StreamRF~\cite{li2022streaming}, a state-of-the-art technique tackling the exactly same task, in all the relevant aspects.

To summarize, our contributions include:

\begin{itemize}
  \item We propose 3DGStream, a method for on-the-fly construction of photo-realistic, real-time renderable FVV on video streams, eliminating the necessity for lengthy offline training on the entire video sequences.

  \item We utilize NTC for modeling the transformations of 3DGs, in conjunction with an adaptive 3DG addition strategy to tackle emerging objects within dynamic scenes. This combination permits meticulous manipulation of 3DGs, accommodating scene alterations with limited performance overhead.

  \item We conduct extensive experiments to demonstrate 3DGStream's competitive edge in rendering quality, training time, and requisite storage, as well as its superior rendering speed, compared to existing state-of-the-art dynamic scene reconstruction methods.
\end{itemize}

%% file: 3DGStream/figs_tabs/teaser_1.tex
\begin{figure}
    \centering
    \begin{subfigure}{0.235\textwidth}
      \includegraphics[width=\textwidth]{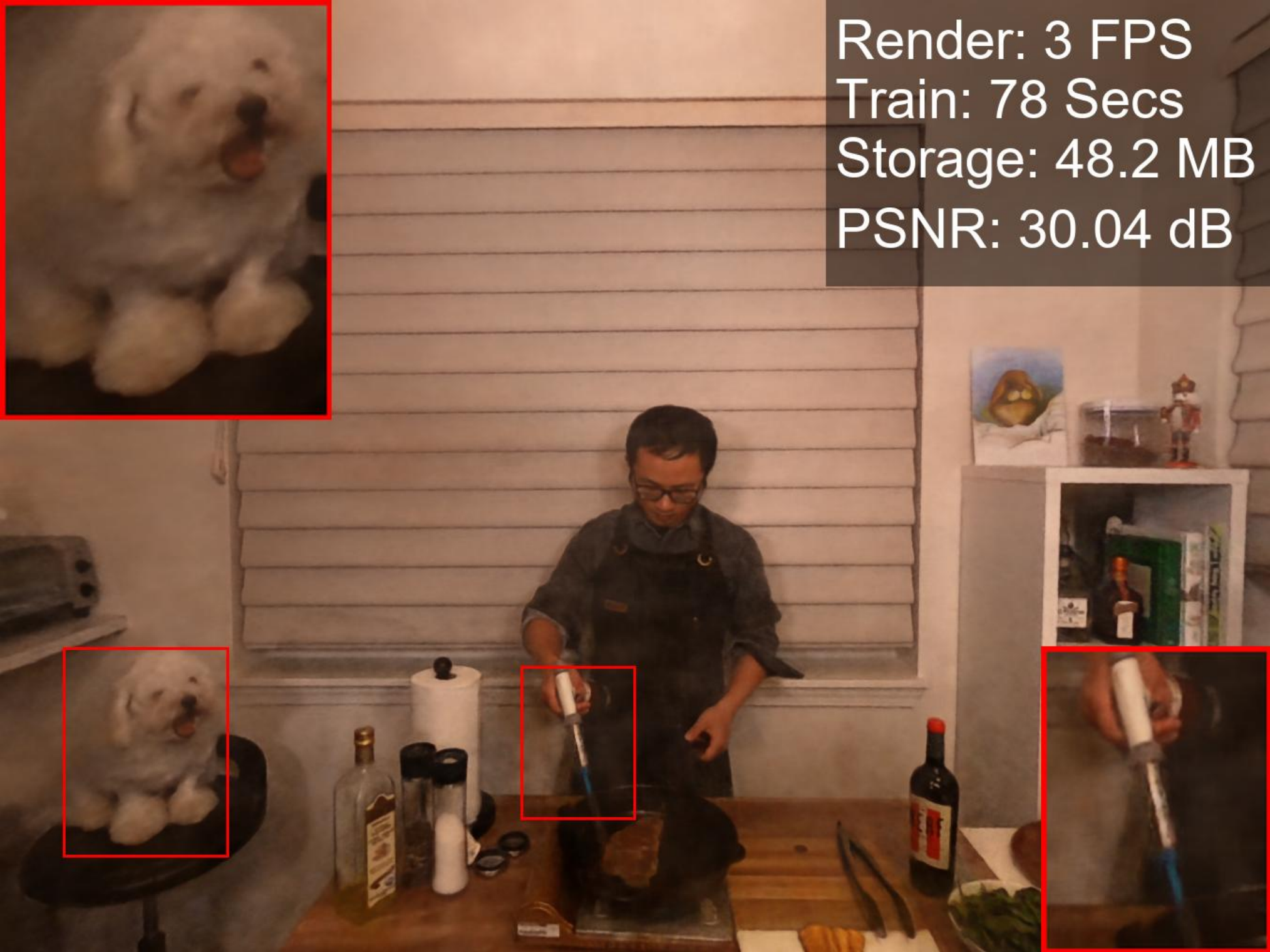}
        \caption*{(a) \textbf{I-NGP}~\cite{muller2022instant}: Per-frame training}
    \end{subfigure}
    \hfill
    \begin{subfigure}{0.235\textwidth}
        \includegraphics[width=\textwidth]{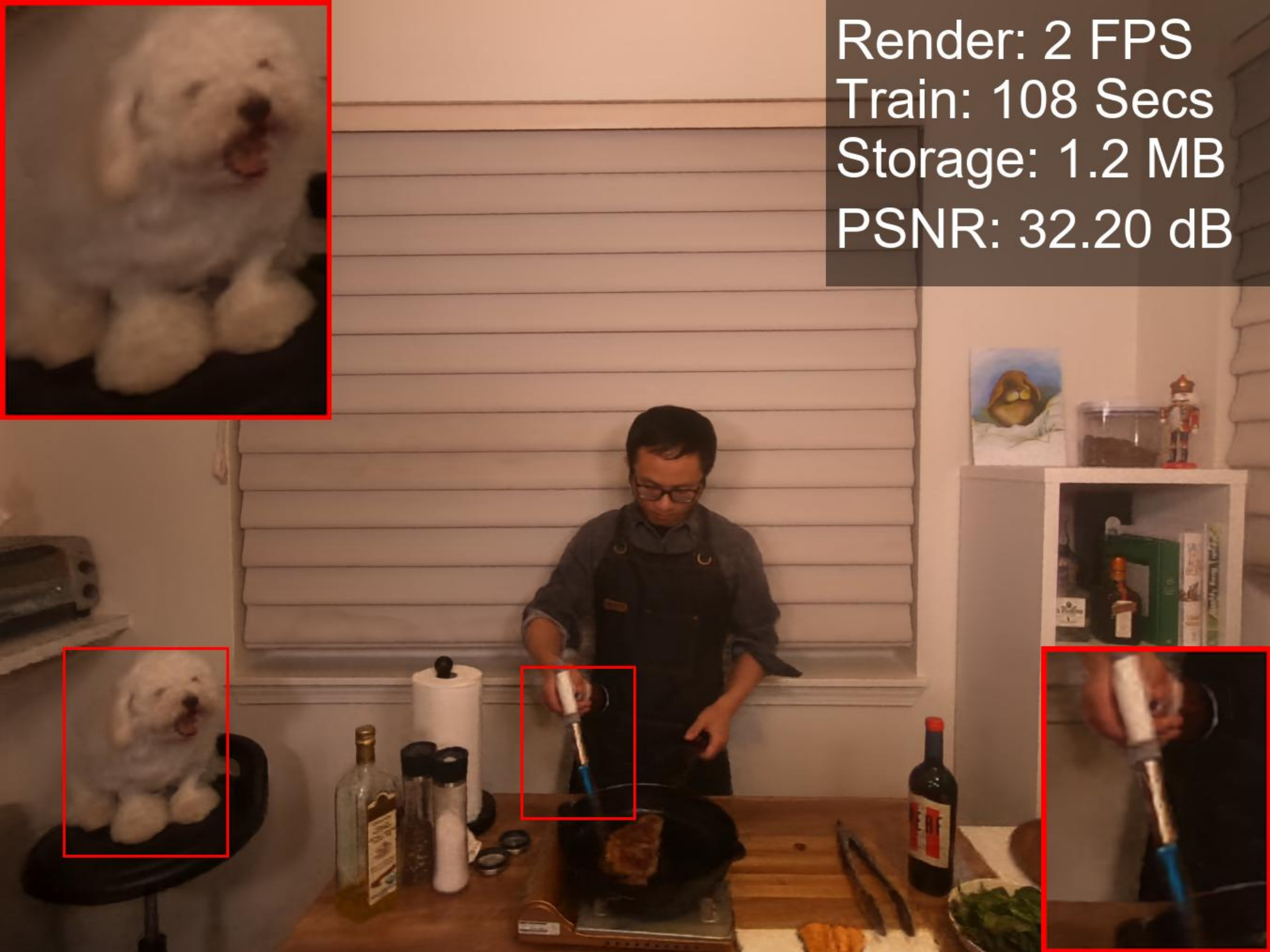}
        \caption*{(b) \textbf{HyperReel}~\cite{attal2023hyperreel}: Offline training}
    \end{subfigure}
    \vspace{0.05cm}    
    \begin{subfigure}{0.235\textwidth}
      \includegraphics[width=\textwidth]{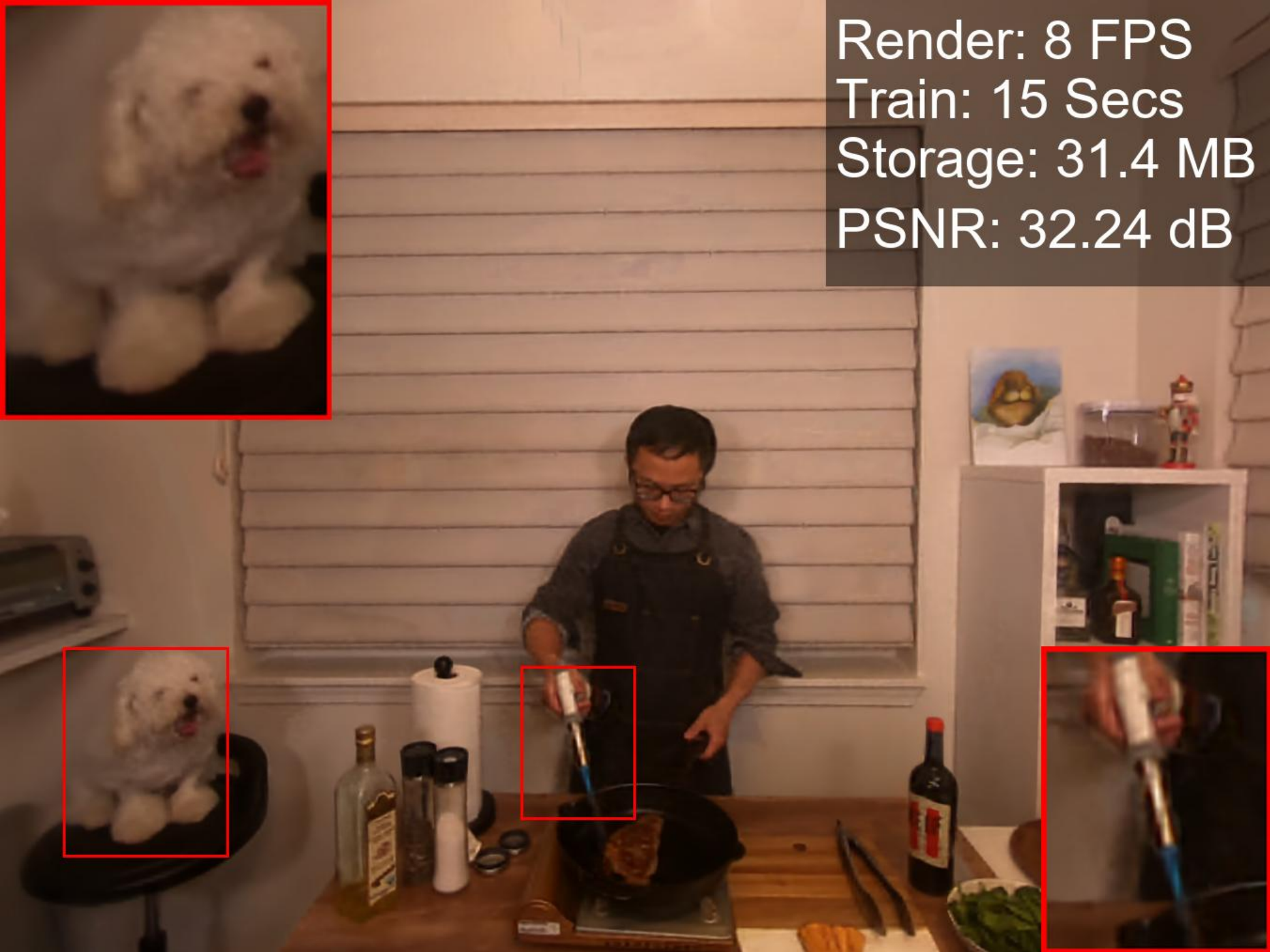}
      \caption*{(c) \textbf{StreamRF}~\cite{li2022streaming}: Online training}
    \end{subfigure}
    \hfill
    \begin{subfigure}{0.235\textwidth}
        \includegraphics[width=\textwidth]{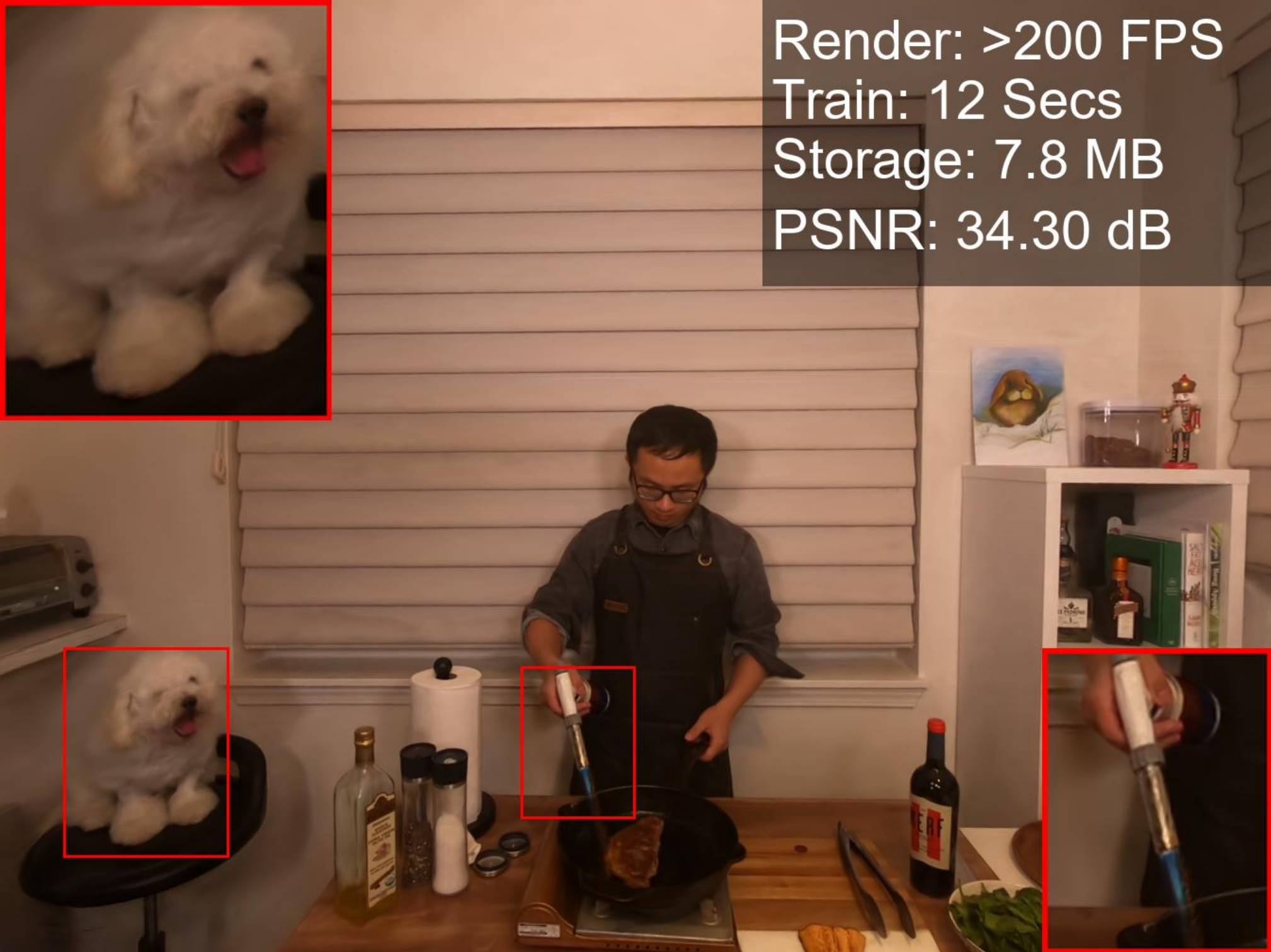}
        \caption*{(d) \textbf{Ours}: Online training}
    \end{subfigure}
    \caption{\textbf{Comparison on the \textit{flame steak} scene of the N3DV dataset~\cite{li2022dynerf}.}
    The training time, requisite storage, and PSNR are computed as averages over the whole video.
    Our method stands out by the ability of fast online training and real-time rendering, standing competitive in both model storage and image quality.}
    \label{fig:teaser_1}
    \vspace{-5mm}
\end{figure}

%% file: 3DGStream/figs_tabs/teaser_2.tex
\begin{figure}
    \centering
    \includegraphics[width=\linewidth]{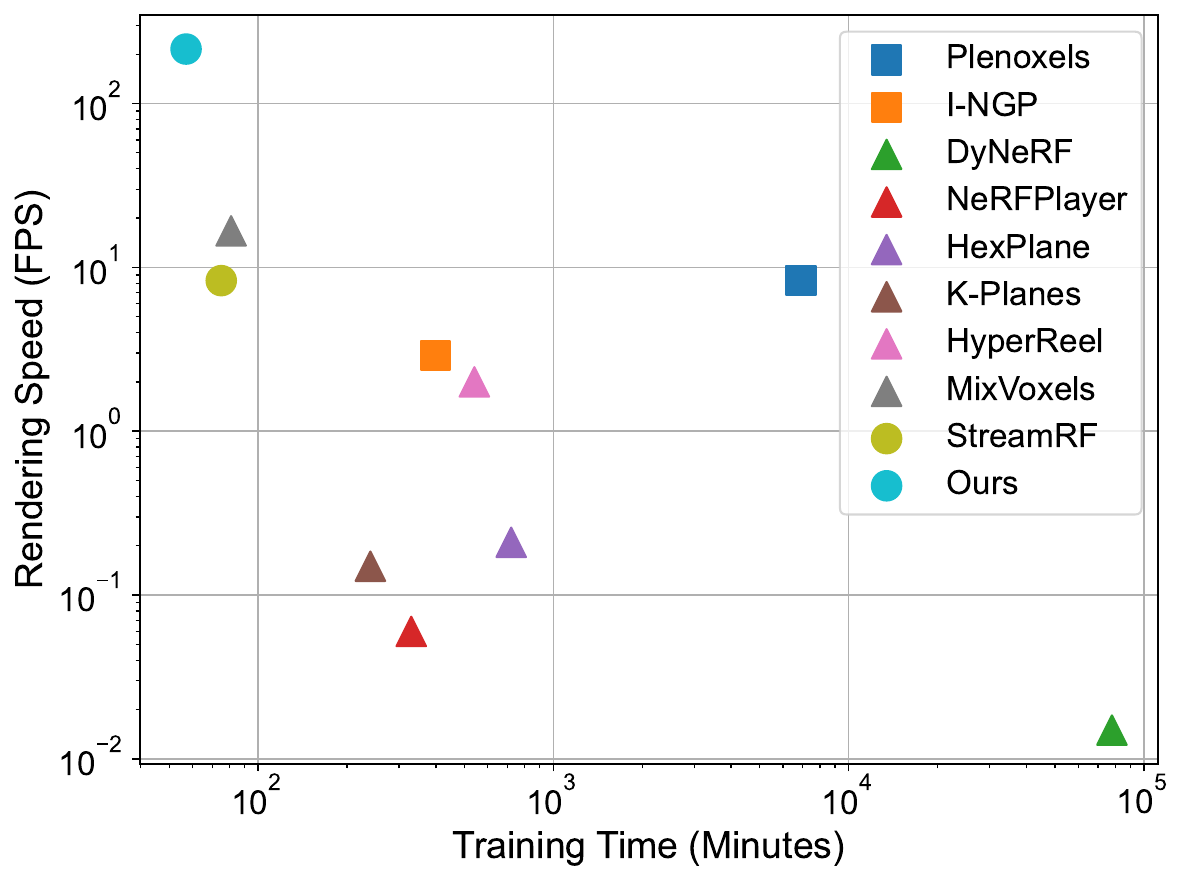}
    \caption{\textbf{Comparison of our method with other methods on the N3DV dataset~\cite{li2022dynerf}.}
    $\Box$ denotes training from scratch per frame, $\bigtriangleup$ represents offline training on complete video sequences, and $\bigcirc$ signifies online training on video streams.
    While achieving online training , our method reaches state-of-the-art performance in both rendering speed and overall training time.}
    \vspace{-5mm}
\label{fig:teaser_2}
\end{figure}

%% file: 3DGStream/2_related_work.tex
\input{3DGStream/figs_tabs/overview}
\section{Related Work}
\label{sec:relative_work}
\subsection{Novel View Synthesis for Static Scenes}
Synthesizing novel views from a set of images of static scenes is a time-honored problem in the domains of computer vision and graphics.
Traditional methods such as Lumigraph~\cite{gortler1996lumigraph,buehler2001unstructured} or
Light-Field~\cite{levoy1996light,shum1999rendering,chai2000plenoptic,davis2012unstructured} achieve new view synthesis through interpolation.
In recent years, Neural Radiance Fields (NeRF)~\cite{mildenhall2020nerf} has achieved photorealistic synthesizing results by representing the radiance field using a multi-layer perceptron (MLP).
A series of subsequent works enhance NeRF's performance in various aspects,
such as accelerating training speeds~\cite{sun2021direct,muller2022instant,chen2022tensorf,fridovich2022plenoxels,hu2023Tri-MipRF,Chen2023TOG},
achieving real-time rendering~\cite{wizadwongsa2021nex,garbin2021fastnerf,chen2022mobilenerf,yu2021plenoctrees,hedman2021snerg,reiser2023merf},
and improving synthesis quality on challenging scenes~\cite{martinbrualla2020nerfw,barron2021mip,barron2022mip,verbin2022refnerf,mildenhall2022nerf,barron2023zipnerf}
or sparse inputs~\cite{yu2020pixelnerf,chen2021mvsnerf,niemeyer2022regnerf,yang2023freenerf,wimbauer2023behind,wynn2023diffusionerf,sun2023vgos}.
Since the vanilla NeRF employs costly volume rendering,
necessitating neural network queries for rendering,
subsequent approaches faced trade-offs in training time,
rendering speed, model storage, image quality, and applicability.
To address these challenges, Kerbl~\etal~\cite{kerbl3Dgaussians} propose 3D Gaussian Splatting (3DG-S), which integrates of 3DGs with differentiable point-based rendering.
3DG-S enables real-time high-fidelity view synthesis in large-scale unbounded scenes after brief training periods with modest storage requirements.
Inspired by this work, we extend its application to the task of constructing FVVs of dynamic scenes.
Taking it a step further, we design a on-the-fly training framework to achieve efficient FVV streaming.
\subsection{Free-Viewpoint Videos of Dynamic Scenes}
\label{sec:relative_work_dynamic}
Constructing FVVs from a set of videos of dynamic scenes is a more challenging and applicable task in the domains of computer vision and graphics.
Earlier attempts to address this task pivoted around the construction of dynamic primitives~\cite{collet2015high,motion2fusion} or resorting to interpolation~\cite{zitnick2004high, broxton2020immersive}.
With the success of NeRF-like methods in novel view synthesis for static scenes,
a series of works~\cite{weng2022humannerf,zhao2022humannerf,li2022tava,yang2022banmo,li2020neural,li2023dynibar,tretschk2020non,park2020deformable,park2021hypernerf,pumarola2021d,song2022nerfplayer,attal2023hyperreel,wang2023omnimotion,li2022dynerf,TiNeuVox,kplanes_2023,Cao2022FWD,park2023temporal,wang2023mixed,li2022streaming,Wang_2023_CVPR,luiten2023dynamic} attempt to use NeRF for constructing FVVs in dynamic scenes.
These works can typically be categorized into five types: prior-driven, flow-based, warp-based, those using spatio-temporal inputs, and per-frame training.

\textbf{Prior-driven} methods~\cite{weng2022humannerf,zhao2022humannerf,li2022tava,yang2022banmo,kirschstein2023nersemble} leverage parametric models or incorporate additional priors, such as skeletons, to bolster performance on the reconstruction of specific dynamic objects, \eg,~humans. However, their application is limited and not generalizable to broader scenes.

\textbf{Flow-based} methods~\cite{li2020neural,li2023dynibar} primarily focus on constructing FVVs from monocular videos. By estimating the correspondence of 3D points in consecutive frames, they achieve impressive results. Nonetheless, the intrinsic ill-posedness of monocular reconstructions in intricate dynamic scenes frequently calls for supplementary priors like depth, optical flow, and motion segmentation masks.

\textbf{Warp-based} methods~\cite{tretschk2020non,park2020deformable,park2021hypernerf,pumarola2021d,song2022nerfplayer,attal2023hyperreel,wang2023omnimotion}
assumpt that the dynamics of the scene arise from the deformation of static structures.
These methods warp the radiance field of each frame onto one or several canonical frames, achieving notable results. However, the strong assumptions they rely on often prevent them from handling topological variations.

Methods that use \textbf{spatio-temporal inputs}~\cite{li2022dynerf,wang2022fourier,TiNeuVox,kplanes_2023,Cao2022FWD,park2023temporal,wang2023mixed}
enhance radiance fields by adding a temporal dimension,
enabling the querying of the radiance field using spatio-temporal coordinates.
While these techniques showcase a remarkable ability to synthesize new viewpoints in dynamic scenes,
the entangled scene parameters can constrain their adaptability for downstream applications.

\textbf{Per-frame training} methods~\cite{li2022streaming,Wang_2023_CVPR,luiten2023dynamic}
adapt to changes in the scene online by leveraging per-frame training, a paradigm we have also adopted.
To be specific, StreamRF~\cite{li2022streaming} employs Plenoxels~\cite{fridovich2022plenoxels} for scene representation and achieves rapid on-the-fly training with minimal storage requirements through techniques like narrow band tuning and difference-based compression.
ReRF~\cite{Wang_2023_CVPR} uses DVGO~\cite{sun2021direct} for scene representation and optimize motion grid and residual grid frame by frame to model inter-frame discrepancies, enabling high-quality FVV streaming and rendering.
Dynamic3DG~\cite{luiten2023dynamic} optimizes simplified 3DGs and integrates physically-based priors for high-quality novel view synthesis on dynamic scenes.

Among the aforementioned works,
only NeRFPlayer~\cite{song2022nerfplayer},
ReRF~\cite{Wang_2023_CVPR},
StreamRF~\cite{li2022streaming},
and Dynamic3DG~\cite{luiten2023dynamic} are able to stream FVVs.
NeRFPlayer achieves FVV streaming through a decomposition module and a feature streaming module, but it is only able to stream pre-trained models.
ReRF and Dynamic3DG are limited to processing scenes with few objects and foreground mask, necessitating minute-level per-frame training times.
StreamRF stands out by requiring only a few seconds for each frame's training to construct high-fidelity FVVs on challenging real-world dynamic scenes with compressed model storage.
However, it falls short in rendering speed.
Contrarily, our approach matches or surpasses StreamRF in training speed, model storage, and image quality, all while achieving real-time rendering at 200 FPS.

\subsection{Concurrent Works}
\label{sec:relative_work_concurrent}
Except for Dynamic3DG, several concurrent works have extended 3DG-S to represent dynamic scenes. 
Deformable3DG~\cite{yang2023deformable3dgs} employs an MLP to model the deformation of 3DGs, while~\cite{wu20234dgaussians} introduces a hexplane-based encoder to enhance the efficency of deformation query. 
Meanwhile, \cite{yang2023gs4d, duan20244d} lift 3DG to 4DG primitives for dynamic scene representation. 
However, these approaches are limited to offline reconstruction and lack streamable capabilities, whereas our work aims to achieve efficient streaming of FVVs with an online training paradigm.

\vspace{-2mm}

%% file: 3DGStream/figs_tabs/overview.tex
\begin{figure*}
    \centering
    \includegraphics[width=\linewidth]{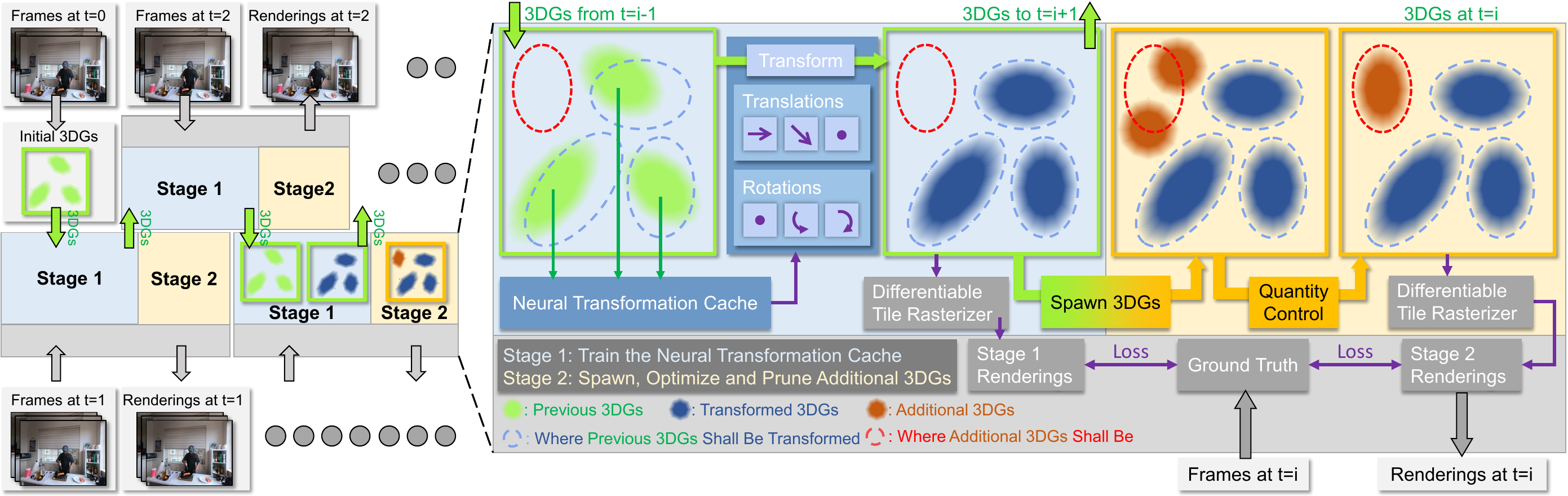}
    \caption{\textbf{Overview of 3DGStream.} Given a set of multi-view video streams, 3DGStream aims to construct high-quality FVV stream of the captured dynamic scene on-the-fly. Initially, we optimize a set of 3DGs to represent the scene at timestep $0$. For each subsequent timestep $i$, we use the 3DGs from timestep $i-1$ as an initialization and then engage in a two-stage training process:
\textbf{Stage 1}: We train the Neural Transformation Cache (NTC) to model the translations and rotations of 3DGs. After training, the NTC transforms the 3DGs, preparing them for the next timestep and the next stage in the current timestep.
\textbf{Stage 2}: We spawn frame-specific additional 3DGs at potential locations and optimize them along with periodic splitting and pruning.
After the two-stage process concludes, both transformed and additional 3DGs are used to render at the current timestep i, with only the transformed ones carried into the next timestep.
}
\label{fig:overview}
\vspace{-5mm}
\end{figure*}

%% file: 3DGStream/3_background.tex
\section{Background: 3D Gaussian Splatting}
\label{sec:background}
3D Gaussian Splatting~(3DG-S)~\cite{kerbl3Dgaussians} employs anisotropic 3D Gaussians as an explicit scene representation.
Paired with a fast differentiable rasterizer, 3DGs achieves real-time novel view synthesis with only minutes of training.
\subsection{3D Gaussians as Scene Representation}
\label{sec:3DGs}
A 3DG is defined by a covariance matrix $\Sigma$ centered at point (\ie,~mean) $\mu$:
\begin{equation}
\label{equ:3DG}
G(x;\mu, \Sigma) = e^{-\frac{1}{2} (x-\mu)^T \Sigma^{-1} (x-\mu)}.
\end{equation}
To ensure positive semi-definiteness during optimization, the covariance matrix $\Sigma$ is decomposed into a rotation matrix $R$ and a scaling matrix $S$:
\begin{equation}
\label{equ:sigma}
\Sigma=RSS^TR^T.
\end{equation}
Rotation is conveniently represented by a unit quaternion,
while scaling uses a 3D vector. Additionally, each 3DG contains a set of spherical harmonics (SH) coefficients of to represent view-dependent colors,
along with an opacity value $\alpha$, which is used in $\alpha$-blending~(\cref{equ:a-blending}).
\subsection{Splatting for Differentiable Rasterization}
\label{sec:render}
For novel view synthesis, 3DG-S~\cite{kerbl3Dgaussians} project 3DGs to 2D Gaussian (2DG) splats~\cite{zwicker2001ewa}:
\begin{equation}
\label{equ:project}
\Sigma' = J W \Sigma W^T J^T.
\end{equation}
Here, $\Sigma'$ is the covariance matrix in camera coordinate. $J$ is the Jacobian of the affine approximation of the projective transformation, and $W$ is the viewing transformation matrix. By skipping the third row and third column of $\Sigma'$, we can derive a $2 \times 2$ matrix denoted as $\Sigma_{2d}$. Furthermore, projecting the 3DG's mean, $\mu$, into the image space results in a 2D mean, $\mu_{2d}$. Consequently, this allows us to define the 2DG in the image space as $G_{2d}(x; \mu_{2d}, \Sigma_{2d})$.

Using $\Sigma'$, the color $C$ of a pixel can be computed by blending the $N$ ordered points overlapping the pixel:
\begin{equation}
\label{equ:a-blending}
C = \sum_{i \in N} c_i \alpha_i' \prod_{j=1}^{i-1} (1 - \alpha_j').
\end{equation}
Here, $c_i$ denotes the view-dependent color of the $i$-th 3DG. $\alpha_i'$ is determined by multiplying the opacity $\alpha_i$ of the $i$-th 3DG $G$ with the evaluation of the corresponding 2DG $G_{2d}$.

Leveraging a highly-optimized rasterization pipeline coupled with custom CUDA kernels, the training and rendering of 3DG-S are remarkably fast. For instance, for megapixel-scale real-world scenes, just a few minutes of optimization allows 3DGs to achieve photo-realistic visual quality and rendering speeds surpassing 100~FPS.

%% file: 3DGStream/4_method.tex
\section{Method}
\label{sec:method}
3DGStream constructs photo-realistic FVV streams from multi-view video streams on-the-fly using a per-frame training paradigm.
We initiate the process by training 3DGs~\cite{kerbl3Dgaussians} at timestep 0.
For subsequent timesteps, we employ the previous timestep's 3DGs as an initialization and pass them to a two-stage pipeline.
Firstly~(\cref{sec:NTC}), a Neural Transformation Cache~(NTC) is trained to model the transformation for each 3DG.
Once the training is finished, we transform the 3DGs and carry the transformed 3DGs to the next timestep.
Secondly~(\cref{sec:ADC}), we employ an adaptive 3DG addition strategy to handle emerging objects.
For each FVV frame, we render views at the current timestep using both the transformed 3DGs and additional 3DGs, while the latter are not passed to the next timestep.
Note that we only need to train and store the parameters of the NTC and the additional 3DGs for each subsequent timestep, not all the 3DGs.
We depict an overview of our approach in~\cref{fig:overview}.
\subsection{Neural Transformation Cache}
\label{sec:NTC}
For NTC, we seek a structure that is compact, efficient, and adaptive to model the transformations of 3DGs.
Compactness is essential to reduce the model storage.
Efficiency enhances training and inference speeds.
Adaptivity ensures the model focuses more on dynamic regions.
Additionally, it would be beneficial if the structure could consider certain priors of dynamic scenes \cite{horn1981determining, black1996robust, tomasi1992shape}, such as the tendency for neighboring parts of an object to have similar motion.

Inspired by Neural Radiance Caching~\cite{muller2021real}
and I-NGP~\cite{muller2022instant},
we employ multi-resolution hash encoding combined with
a shallow fully-fused MLP~\cite{tiny-cuda-nn} as the NTC.
Specifically, following I-NGP, we use multi-resolution voxel grids to represent the scene. Voxel grids at each resolution are mapped to a hash table storing a $d$-dimensional learnable feature vector.
For a given 3D position $x\in \mathbb{R}^3$, its hash encoding at resolution $l$, denoted as $h(x;l)\in \mathbb{R}^d$, is the linear interpolation of the feature vectors corresponding to the eight corners of the surrounding grid.
Consequently, its multi-resolution hash encoding $h(x)=[h(x;0),h(x;1),...,h(x;L-1)]\in\mathbb{R}^{Ld}$, where $L$ represents the number of resolution levels.
The multi-resolution hash encoding addresses all our requirements for the NTC:
\begin{itemize}
\item \textbf{Compactness}: Hashing effectively reduces the storage space needed for encoding the whole scene.
\item \textbf{Efficiency}: Hash table lookup operates in $O(1)$, and is highly compatible with modern GPUs.
\item \textbf{Adaptivity}: Hash collisions occur in hash tables at finer resolutions, allowing regions with larger gradients—representing dynamic regions in our context—to drive the optimization.
\item \textbf{Priors}: The combination of linear interpolation and the voxel-grid structure ensures the local smoothness of transformations. Additionally, the multi-resolution approach adeptly merges global and local information.
\end{itemize}
Furthermore, to enhance the NTC's performance with minimal overhead, we utilize a shallow fully-fused MLP~\cite{tiny-cuda-nn}.
This maps the hash encoding to a 7-dimensional output: the first three dimensions indicate the translation of the 3DG;
the remaining dimensions represent the rotation of the 3DG using quaternions.
Given multi-resolution hash encoding coupled with MLP, our NTC is formalized as:
\begin{equation}
\label{equ:NTC}
{d\mu,dq} = MLP(h(\mu)), 
\end{equation}
where $\mu$ denotes the mean of the input 3DG.
We transform the 3DGs based on $d\mu$ and $dq$. Specifically, the following parameters of the transformed 3DGs are given as:
\begin{itemize}
\item \textbf{Mean}: $\mu'= \mu+d\mu$, where $\mu'$ is the new mean and $+$ represents vector addition.
\item \textbf{Rotation}: $q'=norm(q)\times norm(dq)$, where $q'$ is the new roation, $\times$ indicates quaternion multiplication and $norm$ denotes normalization.
\item \textbf{SH Coefficients}: Upon rotating the 3DG, the SH coefficients should also be adjusted to align with the rotation of the 3DG. Leveraging the rotation invariance of SH, we directly employ SH Rotation to update SHs. Please refer to the supplementary materials (Suppl.) for details.
\end{itemize}
In Stage 1, we transform the 3DGs from the previous frame by NTC and then render with them.
The parameters of the NTC is optimized by the loss between the rendered image and the ground truth.
Following 3DG-S~\cite{kerbl3Dgaussians}, the loss function is $L_1$ combined with a D-SSIM term:
\begin{equation}
\label{equ:render_loss}
L=(1-\lambda)L_1+\lambda L_{D-SSIM},    
\end{equation}
where $\lambda=0.2$ in all our experiments. It should be noted that during the training process,
the 3DGs from the previous frame remain frozen and do not undergo any updates.
This implies that the input to the NTC remains consistent.

Additionally, to ensure training stability, we initialize the NTC with warm-up parameters.
The loss employed during the warm-up is defined as:
\begin{equation}
\label{equ:warm_up_loss}
L_{warm-up}=||d\mu||_1-cos^2(norm(dq),Q), 
\end{equation}
where $Q$ is the identity quaternion.
The first term uses the $L_1$ norm to ensure the estimated translation approaches zero, while the second term, leveraging cosine similarity,
ensures the estimated rotation approximates no rotation.
However, given the double-covering property of the unit quaternions,
we use the square of the cosine similarity.
For each scene, we execute the warm-up solely after the training at timestep 0, using noise-augmented means of the initial 3DGs as input.
After 3000 iterations of training (roughly 20 seconds), the parameters are stored and used to initialize the NTCs for all the following timesteps.
\input{3DGStream/figs_tabs/visual_comparisons}
\subsection{Adaptive 3DG Addition}
\label{sec:ADC}
Relying solely on 3DGs transformations adequately cover a significant portion of real-world dynamic scenes,
with translations effectively managing occlusions and disappearances in subsequent timesteps.
However, this approach falters when faced with objects not present in the initial frame,
such as transient objects like flames or smoke, and new persistent objects like the liquid poured out of a bottle.
Since 3DG is an unstructured explicit representation, it's essential to add new 3DGs to model these emerging objects.
Considering constraints related to model storage requirements and training complexities,
it's not feasible to generate an extensive number of additional 3DGs nor allow them to be used in subsequent frames, as this would cause 3DGs to accumulate over time.
This necessitates a strategy for swiftly generating a limited number of frame-specific 3DGs to model these emerging objects precisely and thereby enhance the completeness of the scene at the current timestep. 

Firstly, we need to ascertain the locations for the emerging objects.
Inspired by 3DG-S~\cite{kerbl3Dgaussians}, we recognized the view-space positional gradients of 3DGs as a key indicator.
We observed that for emerging objects, the 3DGs in proximity exhibited large view-space positional gradients.
This is attributed to the optimization attempting to `masquerade' the emerging object by transforming the 3DGs.
However, since we prevent the colors of the 3DGs from being updated in Stage 1, this attempt falls short.
Nonetheless, they are still transformed to appropriate positions, with large view-space positional gradients.

Based on the aforementioned observations, we deem it appropriate to introduce additional 3DGs around these high-gradient regions.
Moreover, to exhaustively capture every potential location where new objects might emerge, we adopt an \textbf{adaptive} 3DG spawn strategy.
Specifically, we track view-space positional gradient during the final training epoch of Stage 1.
Once this stage concludes, we select 3DGs that have an average magnitude of view-space position gradients exceeding a relatively low threshold $\tau_{grad}=0.00015$.
For each selected 3DG, the position of the additional 3DG is sampled from
$X \sim \mathcal{N}(\mu, 2\Sigma)$,
where $\mu$ and $\Sigma$ is the mean and the convariance matrix of the selected 3DG.
While we avoid assumptions about the other attributes of the additional 3DGs,
improper initializations of SH coefficients and scaling vectors tend to result in an optimization preference for reducing opacity over adjusting these parameters.
This causes additional 3DGs to quickly become transparent, thereby failing to capture the emerging objects.
To mitigate this issue, the SH coefficients and scaling vectors of these 3DGs are derived from the selected ones,
with rotations set to the identity quaternion q = [1, 0, 0, 0] and opacity initialized at 0.1. After spawning,
the 3DGs undergo optimization utilizing the same loss function (\cref{equ:render_loss}) as Stage 1.
Note that only the parameters of the additional 3DGs are optimized, while those of the transformed 3DGs remain fixed.

To guard against local minima and manage the number of additional 3DGs, we implement an \textbf{adaptive} 3DG quantity control strategy.
Specifically, in Stage 2, we set a relatively high threshold, $\tau_{\alpha}=0.01$, for the opacity value.
At the end of each training epoch, for 3DGs with view-space position gradients exceeding $\tau_{grad}$,
we spawn additional 3DGs nearby to address under-reconstructed regions.
These additional 3DGs inherit their rotations and SH coefficients from the original 3DG,
but their scaling is adjusted to 80\% of the original,
mirroring the `split' operation described by Kerbl et al.~\cite{kerbl3Dgaussians}.
Subsequently, we discard any additional 3DGs with opacity values below $\tau_{\alpha}$ to suppress the growth in the quantity of 3DGs. 


\input{3DGStream/figs_tabs/N3DV_comparisons_avg}
\input{3DGStream/figs_tabs/meetroom_comparisons}

%% file: 3DGStream/figs_tabs/visual_comparisons.tex
\begin{figure*}
    \centering
    \begin{subfigure}{0.195\textwidth}
      \includegraphics[width=\textwidth]{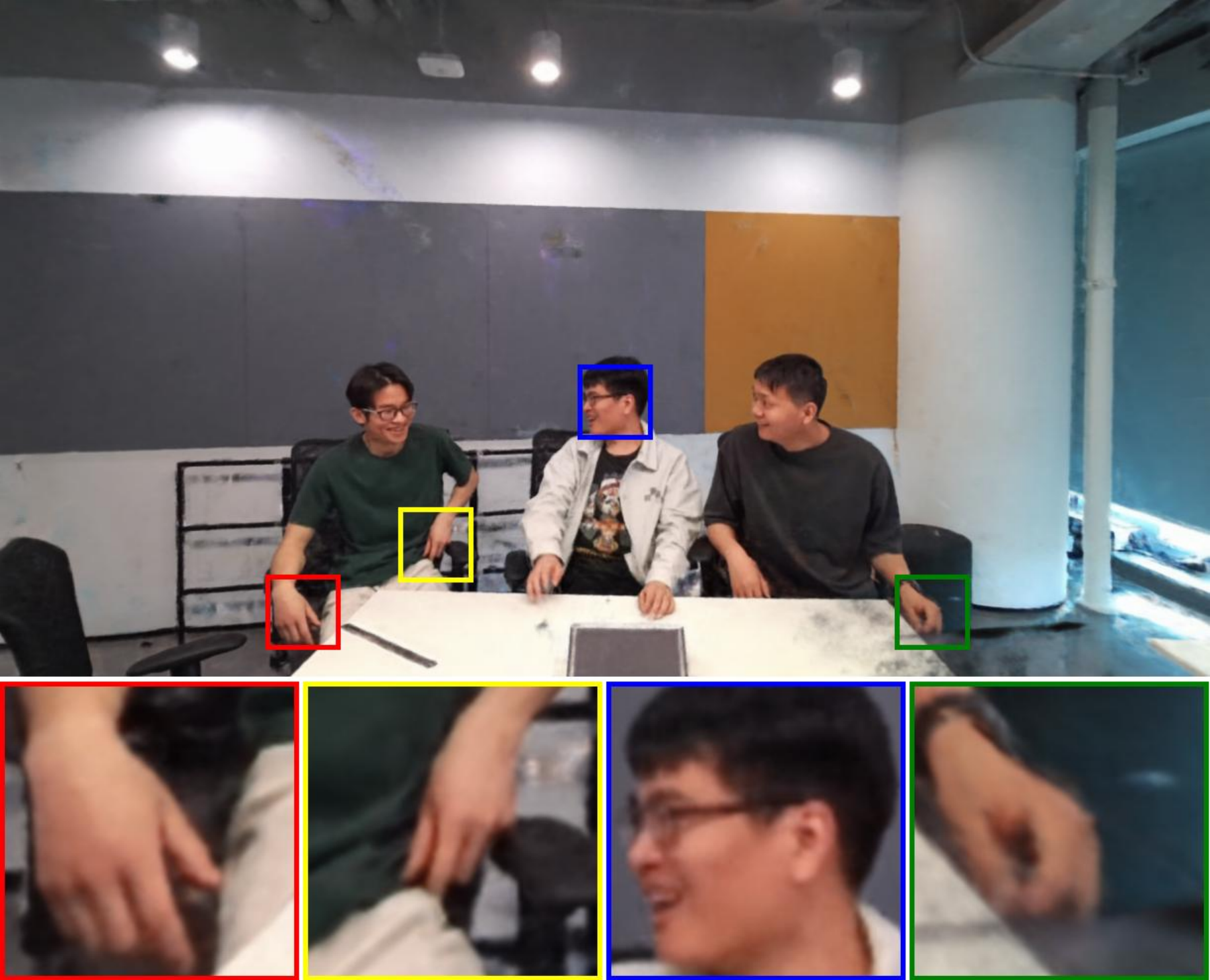}
    \end{subfigure}
    \hfill
    \begin{subfigure}{0.195\textwidth}
        \includegraphics[width=\textwidth]{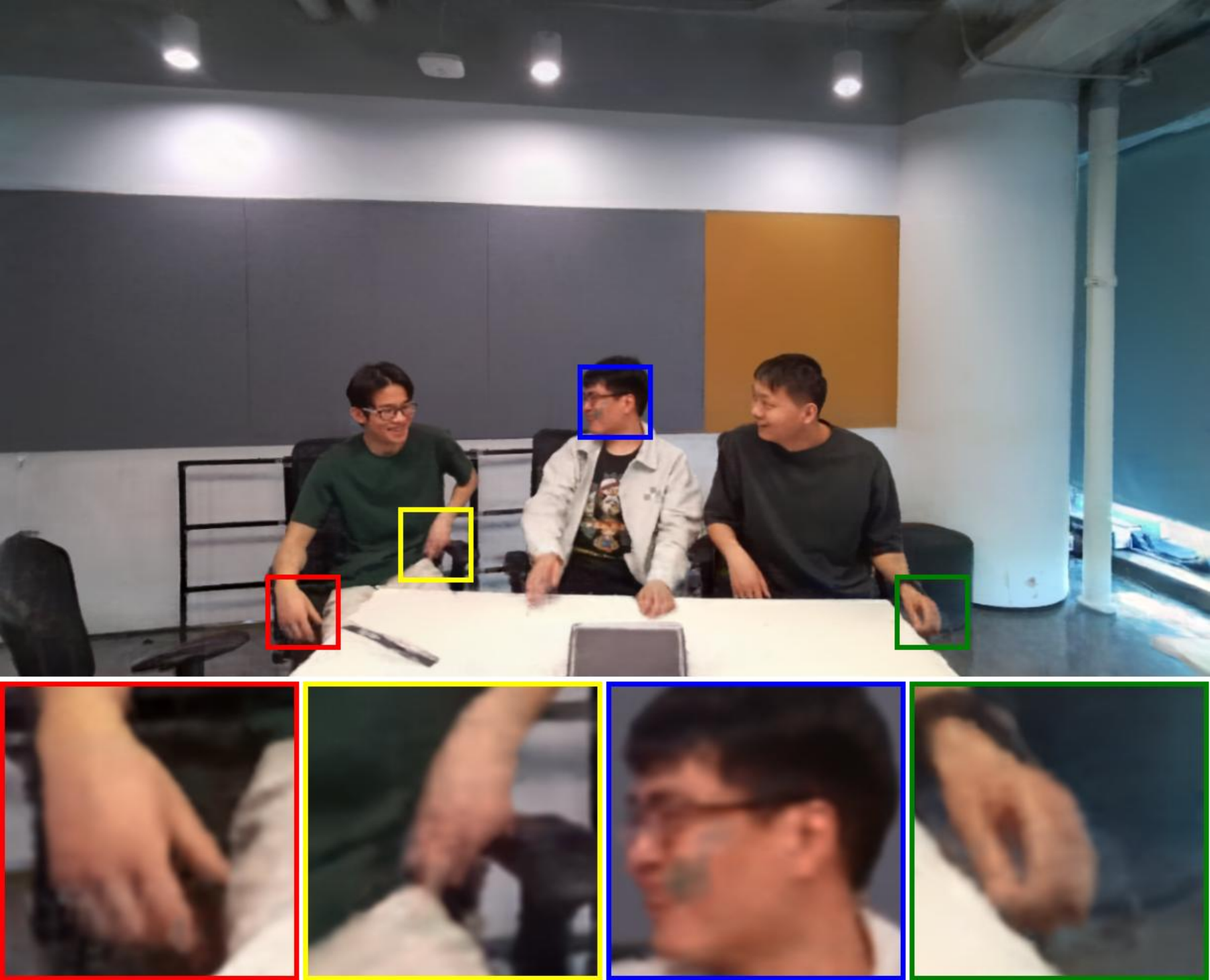}
    \end{subfigure}
    \hfill
    \begin{subfigure}{0.195\textwidth}
        \includegraphics[width=\textwidth]{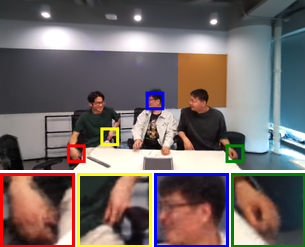}
    \end{subfigure}
    \hfill
    \begin{subfigure}{0.195\textwidth}
      \includegraphics[width=\textwidth]{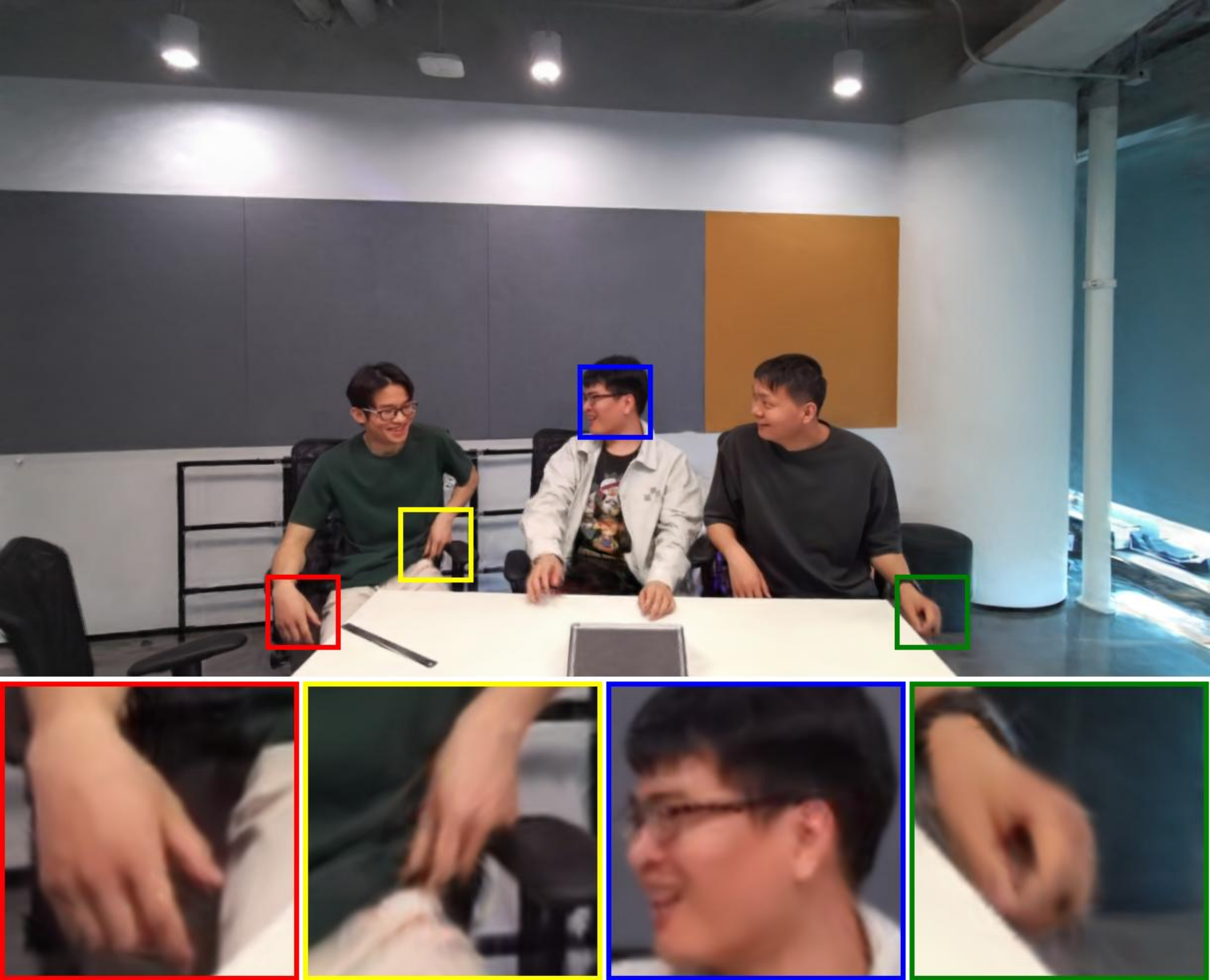}
  
    \end{subfigure}
    \hfill
    \begin{subfigure}{0.195\textwidth}
        \includegraphics[width=\textwidth]{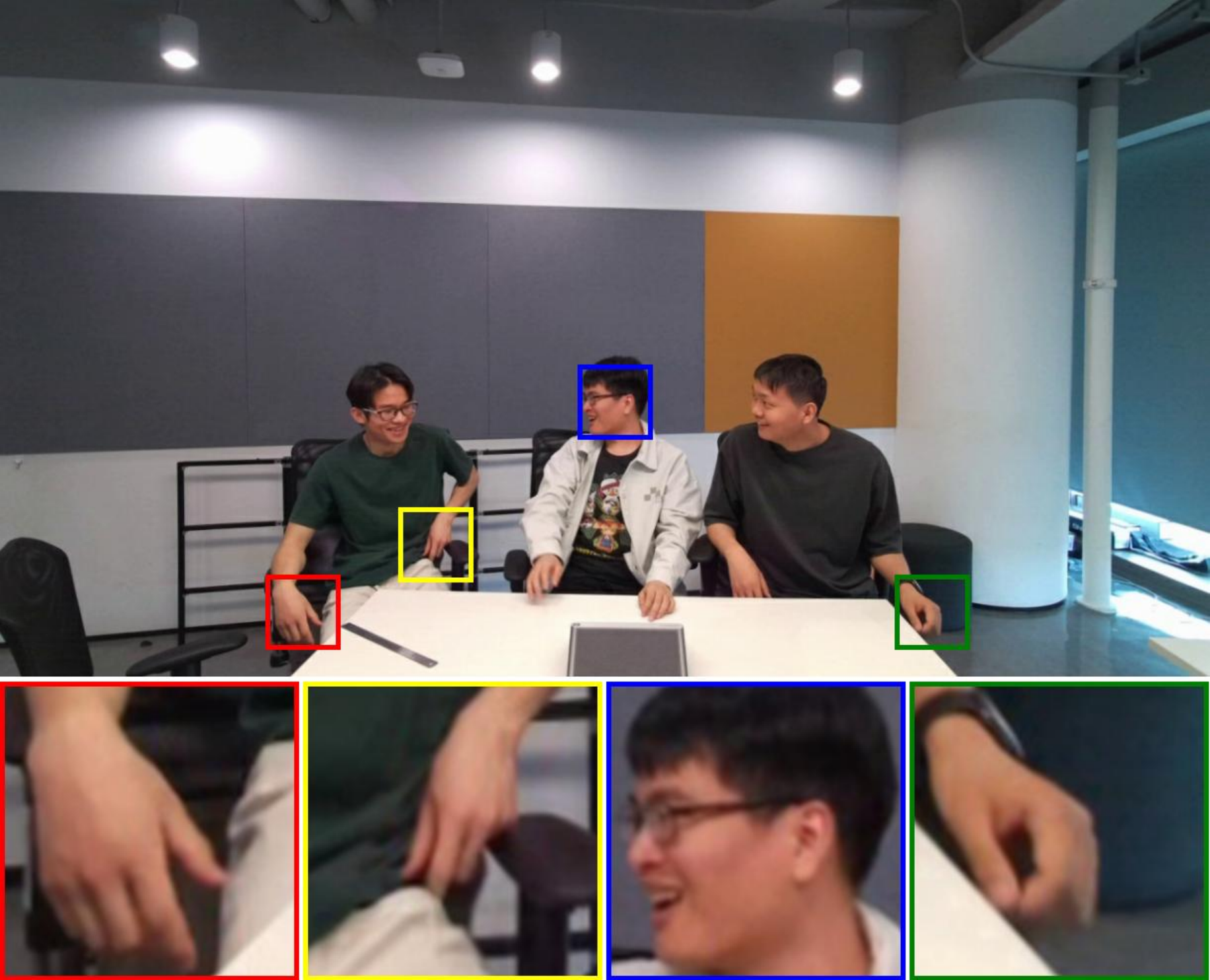}
    \end{subfigure}
    
    \vspace{0.05cm}
    
    \begin{subfigure}{0.195\textwidth}
      \includegraphics[width=\textwidth]{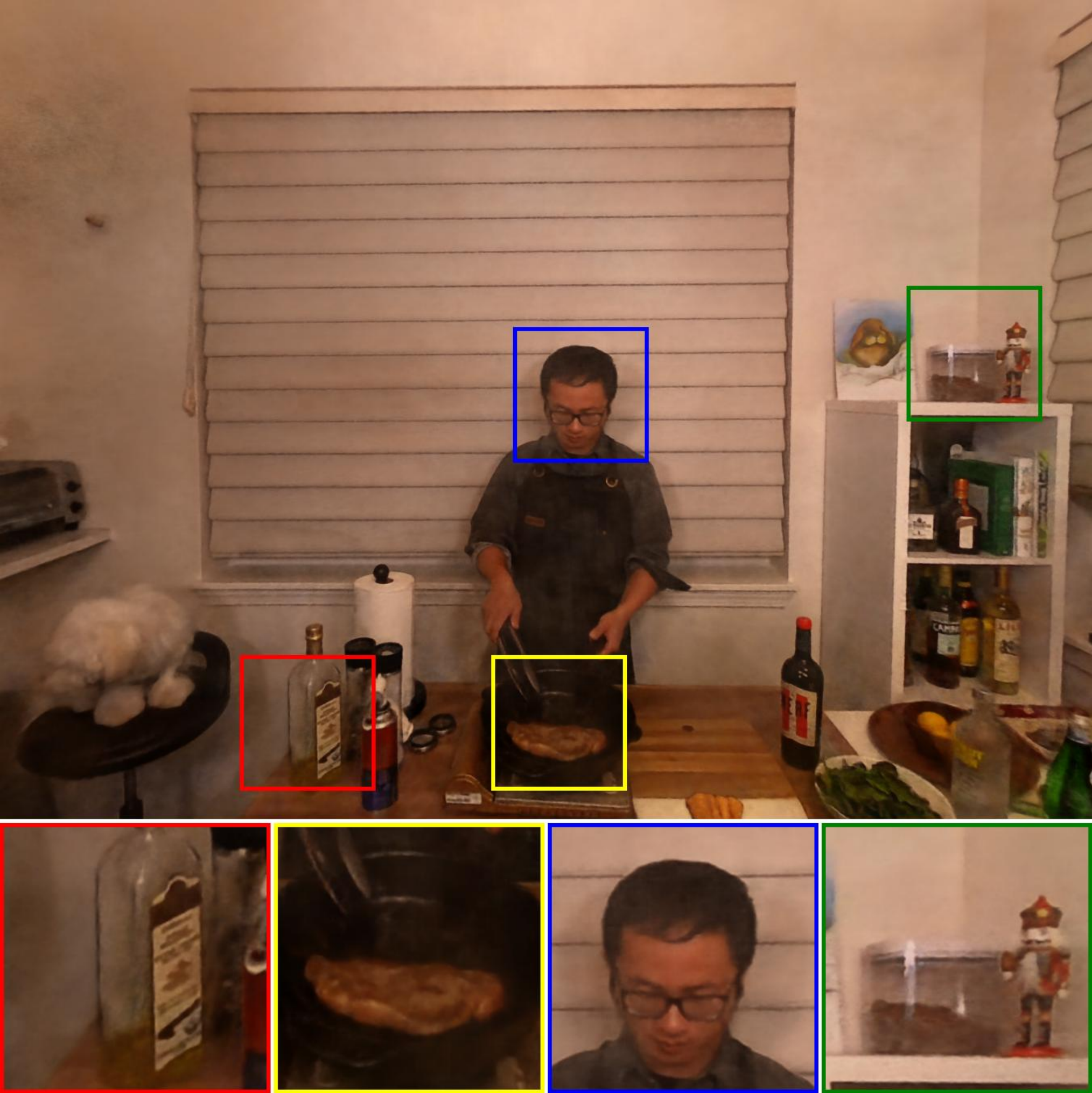}
        \caption*{(a) I-NGP~\cite{muller2022instant}}
    \end{subfigure}
    \hfill
    \begin{subfigure}{0.195\textwidth}
        \includegraphics[width=\textwidth]{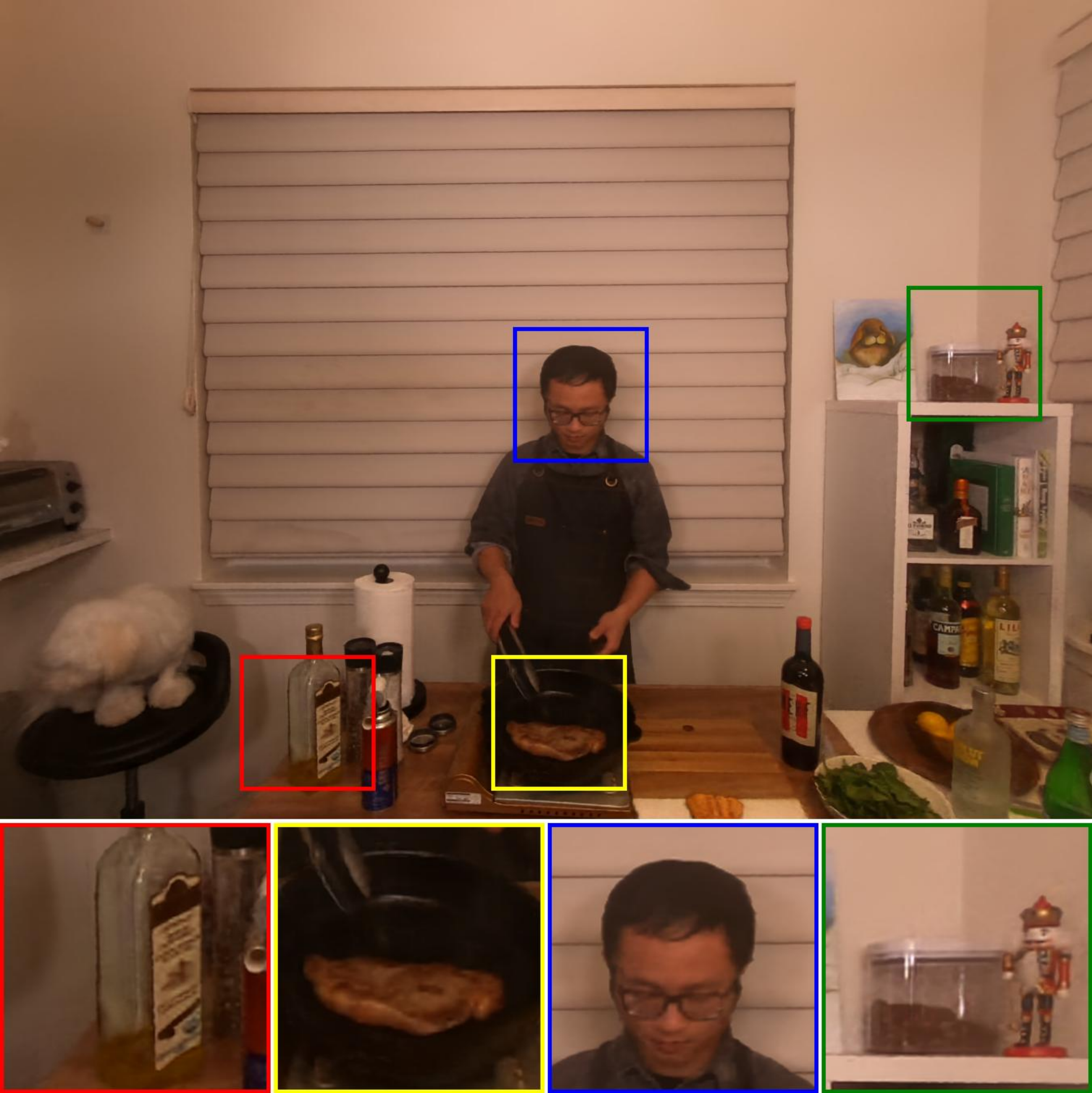}
        \caption*{(b) HyperReel~\cite{attal2023hyperreel}}
    \end{subfigure}
    \hfill
    \begin{subfigure}{0.195\textwidth}
        \includegraphics[width=\textwidth]{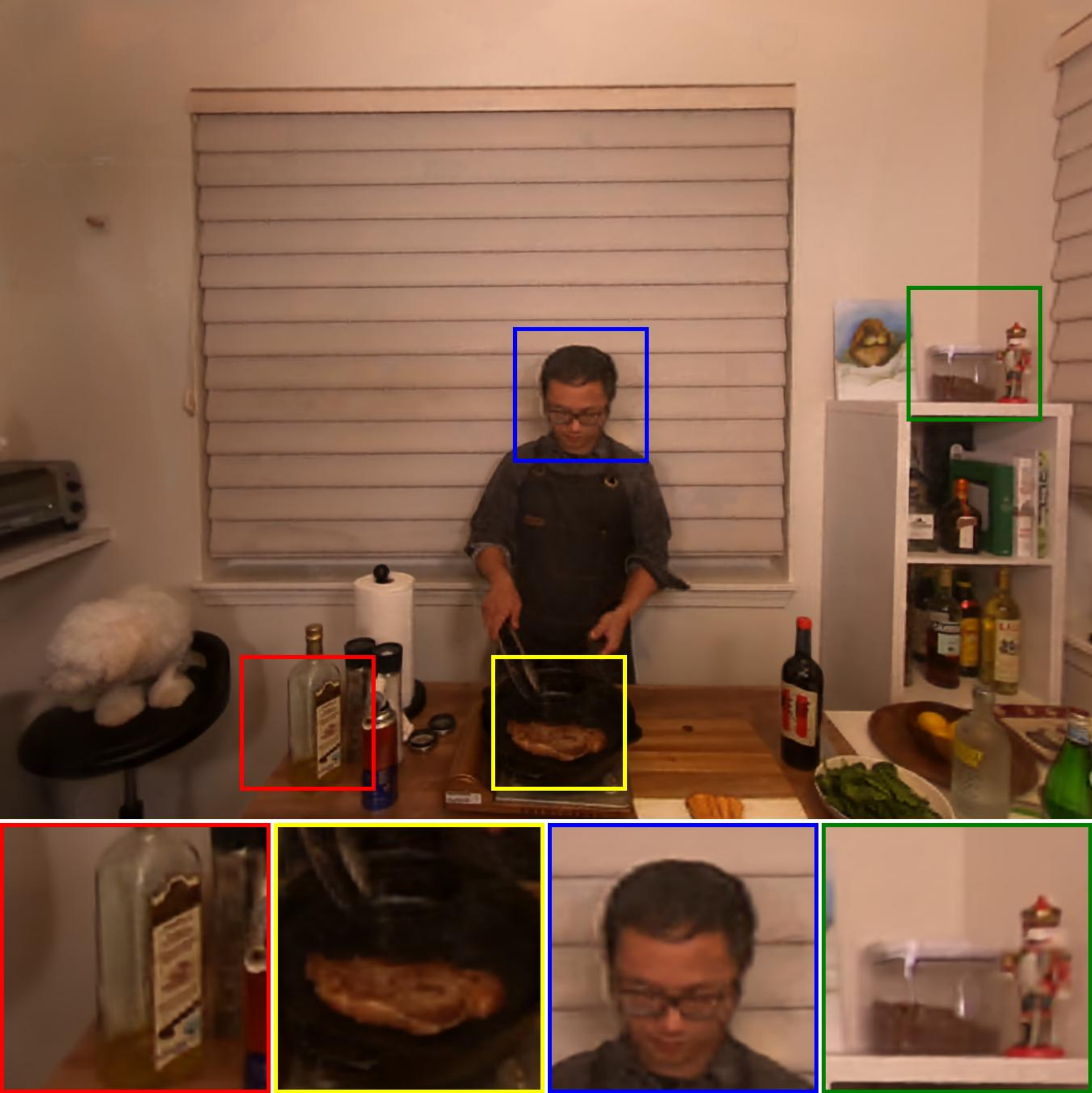}
        \caption*{(c) StreamRF~\cite{li2022streaming}}
    \end{subfigure}
    \hfill
    \begin{subfigure}{0.195\textwidth}
      \includegraphics[width=\textwidth]{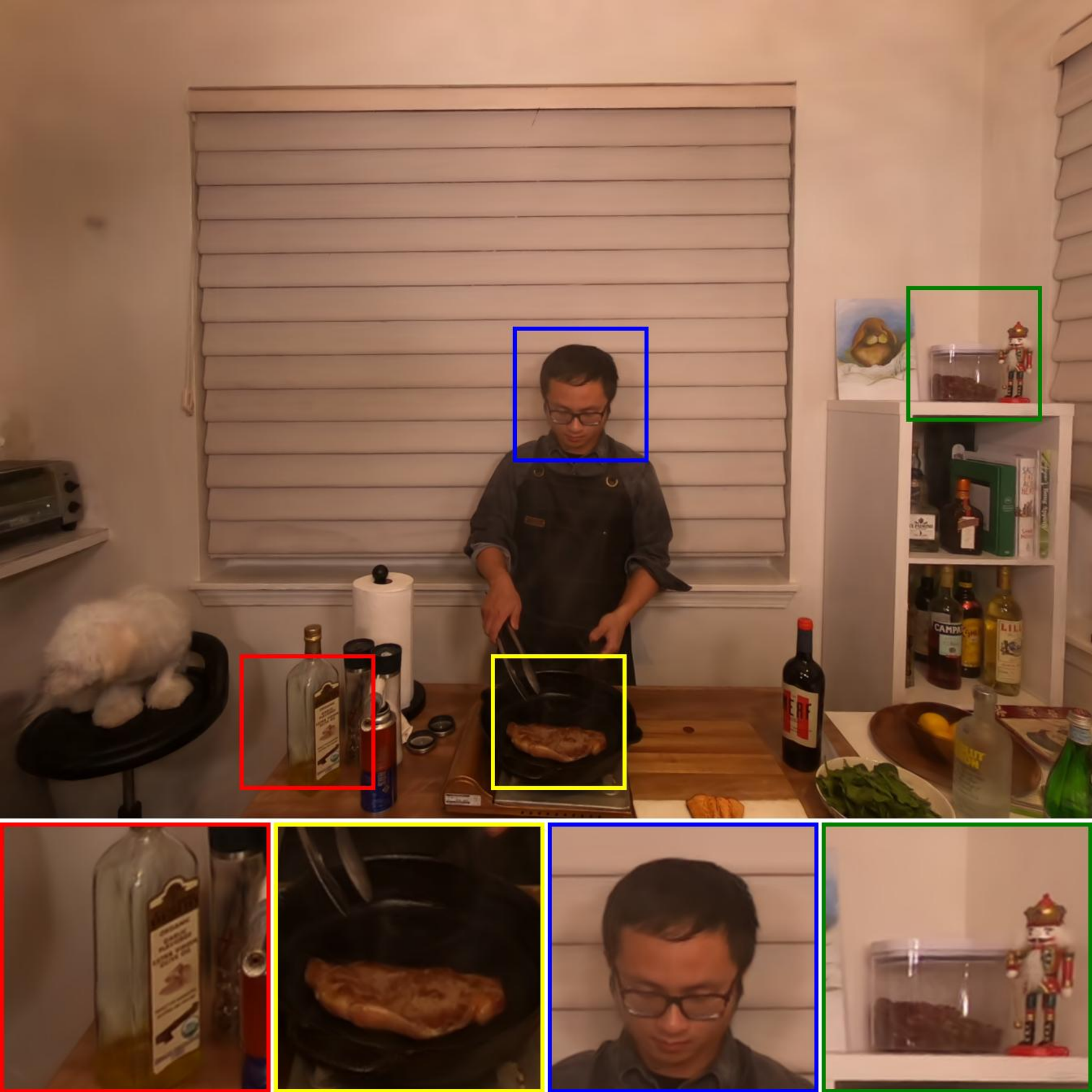}
      \caption*{(d) 3DGStream}
    \end{subfigure}
    \hfill
    \begin{subfigure}{0.195\textwidth}
        \includegraphics[width=\textwidth]{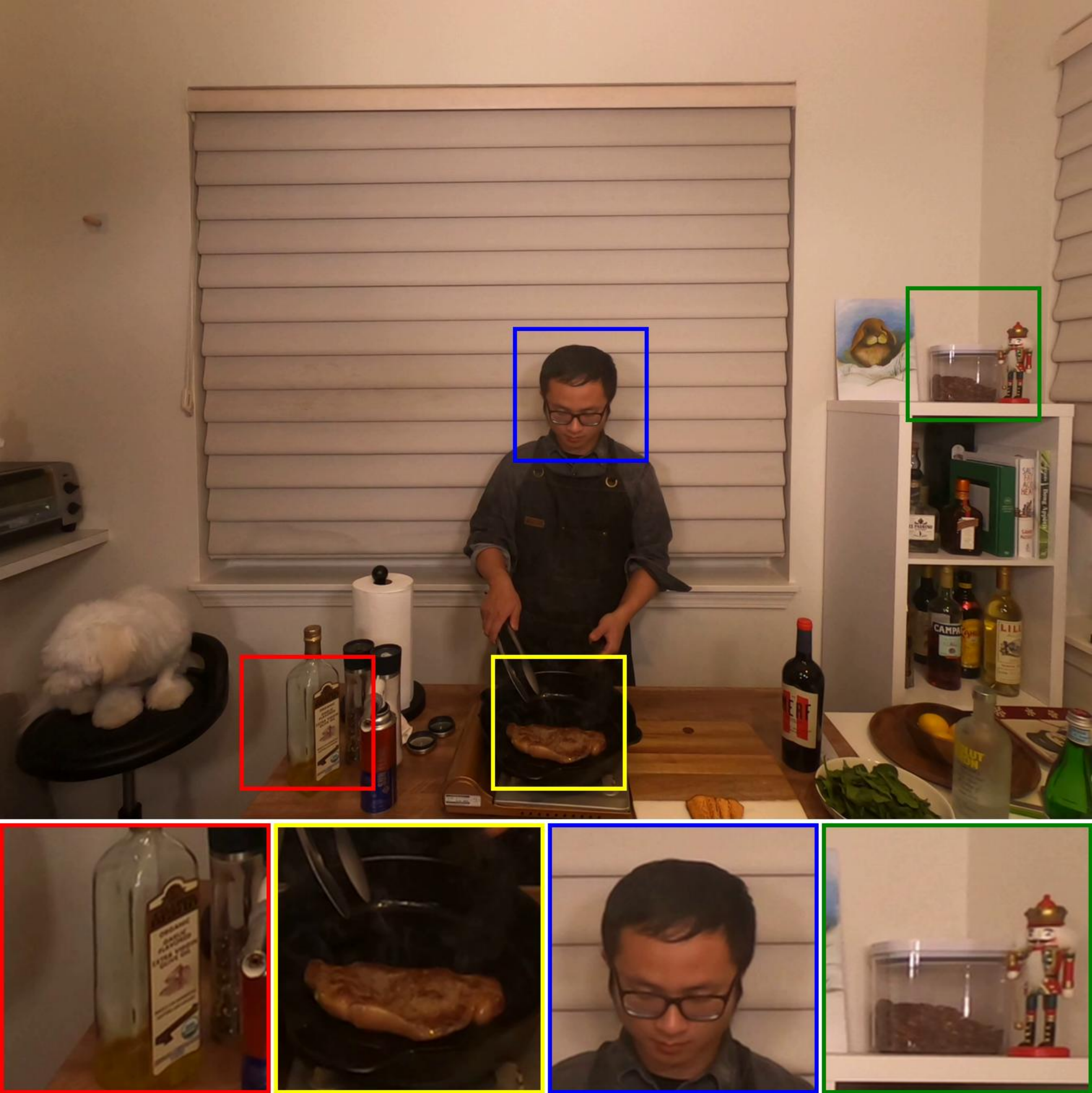}
        \caption*{(e) Ground Truth}
    \end{subfigure}
    \caption{\textbf{Qualitative comparisons} on the \textit{discussion} scene of the Meet Room dataset and the \textit{sear steak} scene of the N3DV dataset.}
    \label{fig:visual_Comparisons}
    \vspace{-5mm}
\end{figure*}

%% file: 3DGStream/figs_tabs/N3DV_comparisons_avg.tex
\begin{table}
    \centering
    \resizebox{\columnwidth}{!}{%
    \begin{tabular}{@{}c|l|cccc|c@{}}
      \toprule
      \multirow{2}{*}{Category} & \multirow{2}{*}{Method}                 & PSNR$\uparrow$& Storage$\downarrow$ & Train$\downarrow$   & Render$\uparrow$       & \multirow{2}{*}{Streamable}\\
                                &                                         & (dB)          & (MB)    & (mins)  & (FPS)        & \\
      \midrule
      \multirow{3}{*}{Static}   & Plenoxels~\cite{fridovich2022plenoxels} & 30.77         & 4106    & 23      & 8.3          & $\checkmark$  \\
                                & I-NGP~\cite{muller2022instant}          & 28.62         & 48.2    & 1.3     & 2.9         & $\checkmark$   \\
                                & 3DG-S~\cite{kerbl3Dgaussians}           & \cellcolor{first}32.08         & 47.1    & 8.3     & \cellcolor{first}390 & $\checkmark$ \\
      \midrule
      \multirow{6}{*}{Offline}  & DyNeRF~\cite{li2022dynerf}    & 29.58$^\dagger$         & \cellcolor{first}0.1 & 260  & 0.02       & $\times$ \\
                                & NeRFPlayer~\cite{song2022nerfplayer}    & 30.69         & 17.1        & 1.2  & 0.05       & $\checkmark$ \\
                                & HexPlane~\cite{Cao2022FWD}              & \cellcolor{second}31.70          & \cellcolor{second}0.8 & 2.4  & 0.21   & $\times$ \\
                                & K-Planes~\cite{kplanes_2023}            & 31.63         & \cellcolor{third}1.0          & 0.8  & 0.15   & $\times$ \\
                                & HyperReel~\cite{attal2023hyperreel}     & 31.10             & 1.2          & 1.8  & 2.00   & $\times$ \\
                                & MixVoxels~\cite{wang2023mixed}          & 30.80             & 1.7          & \cellcolor{third}0.27 & \cellcolor{third}16.7   & $\times$ \\
      \midrule
      \multirow{2}{*}{Online}   & StreamRF~\cite{li2022streaming}         & 30.68             & 17.7/31.4$^\star$  & \cellcolor{second}0.25 & 8.3 & $\checkmark$  \\
                                & Ours                                    & \cellcolor{third}31.67             & 7.6/7.8$^\star$    & \cellcolor{first}0.20    & \cellcolor{second}215 & $\checkmark$ \\
      \bottomrule
    \end{tabular}
    }
    \caption{\textbf{Quantitative comparison} on the N3DV dataset.
    The training time, required storage and PSNR are averaged over the whole 300 frames for each scene.
    $^\dagger$DyNeRF~\cite{li2022dynerf} only report metrics on the \textit{flame salmon} scene.
    $^\star$Considering the initial model.}
    \label{tab:N3DV_Comparisons_Avg}
    \vspace{-4mm}
\end{table}

%% file: 3DGStream/figs_tabs/meetroom_comparisons.tex
\begin{table}
    \centering
    \begin{tabular}{@{}l|cccc@{}}
    \toprule
    \multirow{2}{*}{Method}  & PSNR$\uparrow$& Storage$\downarrow$ & Train$\downarrow$   & Render$\uparrow$\\
                             & (dB)          & (MB)    & (mins)  & (FPS)\\
    \midrule
    Plenoxels~\cite{fridovich2022plenoxels} & 27.15 & 1015   & 14   & 10     \\
    I-NGP~\cite{muller2022instant}      & 28.10 & 48.2   & 1.1    & 4.1   \\
    3DG-S~\cite{kerbl3Dgaussians}     & \cellcolor{first}31.31 & 21.1    & 2.6    & \cellcolor{first}571    \\
    \midrule
    StreamRF~\cite{li2022streaming}  & 26.72 & \cellcolor{second}5.7/9.0$^\star$ & \cellcolor{second}0.17    & 10     \\
    Ours      & \cellcolor{second}30.79 & \cellcolor{first}4.0/4.1$^\star$ & \cellcolor{first}0.10    & \cellcolor{second}288    \\
    \bottomrule
    \end{tabular}
    \caption{\textbf{Quantitative comparison} on the Meet Room dataset.
    Note that the training time, required storage and PSNR are averaged over the whole 300 frames.
    $^\star$Considering the initial model.}
    \label{tab:meetroom_Comparisons}
    \vspace{-5mm}
\end{table}

%% file: 3DGStream/5_evaluation.tex
\section{Experiments}
\subsection{Datasets}
We conduct experiments on two real-world dynamic scene datasets: N3DV dataset~\cite{li2022dynerf} and Meet Room dataset~\cite{li2022streaming}.

\textbf{N3DV dataset}~\cite{li2022dynerf} is captured using a multi-view system of 21 cameras,
comprises dynamic scenes recorded at a resolution of 2704$\times$2028 and 30 FPS.
Following previous works~\cite{li2022dynerf,Cao2022FWD,kplanes_2023,li2022streaming,song2022nerfplayer,wang2023mixed},
we downsample the videos by a factor of two and follow the training and validation camera split provided by~\cite{li2022dynerf}.

\textbf{Meet Room dataset}~\cite{li2022streaming} is captured using a 13-camera multi-view system,
comprises dynamic scenes recorded at a resolution of 1280$\times$720 and 30 FPS.
Following~\cite{li2022streaming}, we utilize 13 views for training and reserved 1 for testing.

\subsection{Implementation}
We implement 3DGStream upon the codes of 3D Gaussian Splatting~(3DG-S)~\cite{kerbl3Dgaussians}, and implement the Neural Transformation Cache (NTC) using tiny-cuda-nn~\cite{tiny-cuda-nn}.
For the training of initial 3DGs, we fine-tune the learning rates on the N3DV dataset based on the default settings of 3DG-S,
and apply them to the Meet Room dataset.
For all scenes, we train the NTC for 150 iterations in Stage 1.
and train the additional 3DGs for 100 iterations in Stage 2.
Please refer to Suppl. for more details.
\input{3DGStream/figs_tabs/NTC_eva}
\input{3DGStream/figs_tabs/s1_eva}
\input{3DGStream/figs_tabs/s2_eva}
\input{3DGStream/figs_tabs/s2_abl}
\subsection{Comparisons}
\textit{Quantitative comparisons.}~Our quantitative analysis involves benchmarking 3DGStream on the N3DV dataset and Meet Room dataset,
comparing it with a range of representative methods.
We take Plenoxels~\cite{fridovich2022plenoxels}, I-NGP~\cite{muller2022instant},
and 3DG-S~\cite{kerbl3Dgaussians} as representatives of fast static scene reconstruction methods,
training them from scratch for each frame.
StreamRF~\cite{li2022streaming}, Dynamic3DG~\cite{luiten2023dynamic}, and ReRF~\cite{wang2023neuralresidual} are designed for online training in dynamic scenes.
Owing to the limitations of Dynamic3DG and ReRF,
which necessitate foreground masks and are confined to scenes with fewer objects,
and their minute-level per-frame training times,
we select StreamRF selected as the representative for online training methods due to its adaptability and training feasibility on the N3DV and MeetRoom datasets.
To demonstrate 3DGStream’s competitive image quality,
we drew comparisons with the quantitative results reported for the N3DV dataset in the respective papers of DyNeRF~\cite{li2022dynerf},
NeRFPlayer~\cite{song2022nerfplayer}, HexPlane~\cite{Cao2022FWD}, K-Planes~\cite{kplanes_2023}, HyperReel~\cite{attal2023hyperreel}, and MixVoxels~\cite{wang2023mixed},
all of which are methods for reconstructing dynamic scenes through offline training on entire video sequences.

In~\cref{tab:N3DV_Comparisons_Avg}, we present the averaged rendering speed, training time, required storage, and peak signal-to-noise ratio (PSNR) over all scenes of the N3DV dataset.
For each scene, the latter three metrics are computed as averages over the whole 300 frames.
Besides, we provide a breakdown of comparisons across all scenes within the N3DV dataset in the Suppl.
To demonstrate the generality of our method, we conducted experiments on the MeetRoom dataset,
as introduced by StreamRF~\cite{li2022streaming},
and performed a quantitative comparison against Plenoxels~\cite{fridovich2022plenoxels},
I-NGP~\cite{muller2022instant},
3DG-S~\cite{kerbl3Dgaussians},
and StreamRF~\cite{li2022streaming}.
The results are presented in~\cref{tab:meetroom_Comparisons}.
As presented in~\cref{tab:N3DV_Comparisons_Avg,tab:meetroom_Comparisons},
our method demonstrates superiority through fast online training and real-time rendering,
concurrently maintaining a competitive edge in terms of model storage and image quality.
Furthermore, among the methods capable of streaming FVVs, our model requires the minimal model storage.

\textit{Qualitative comparisons.}~While our approach primarily aims to enhance the efficiency of online FVV construction,
as illustrated in \cref{tab:N3DV_Comparisons_Avg,tab:meetroom_Comparisons},
it still achieves competitive image quality.
In~\cref{fig:visual_Comparisons}, we present a qualitative comparison with
I-NGP~\cite{muller2022instant},
HyperReel~\cite{attal2023hyperreel},
and StreamRF~\cite{li2022streaming}
across scenes on the N3DV dataset~\cite{li2022dynerf} and the Meet Room dataset~\cite{li2022streaming},
with a special emphasis on dynamic objects such as faces, hands, and tongs, as well as intricate objects like labels and statues.
It is evident that our method faithfully captures the dynamics of the scene without sacrificing the ability to reconstruct intricate objects.
Please refer to our project page for more video results.
\input{3DGStream/figs_tabs/rendering}
\subsection{Evaluations}
\textit{Neural Transformation Cache.}
We utilize distinct approaches to model the transformations of 3DGs from the first to the second frame within the \textit{flame salmon} video of the N3DV dataset to show the effectiveness of NTC.
\cref{fig:NTC_eva} shows that, without multi-resolution hash encoding~(\textit{w/o} Hash enc.), the MLP faces challenges in modeling transformations effectively.
Additionally, without the warm-up~(\textit{w/o} Warm-up), it takes more iterations for convergence.
Besides, even when compared with the direct optimization of the previous frame's 3DGs~(Direct Opt.), NTC demonstrate on-par performance.
In~\cref{fig:s1_eva},
We present the results of different approaches applied across the entire \textit{flame salmon} video, excluding the first frame~(\ie,~Frame 0).
\textit{w/o} Hash enc. and \textit{w/o} Warm-up. are not able to converge swiftly, resulting in accumulating errors as the sequence progresses.
Direct Opt. yields the best outcomes but at the cost of inflated storage.
Utilizing NTC, in contrast, delivers comparable results with substantially lower storage overhead by eliminating the need for saving all the 3DGs.

\textit{Adaptive 3DG Addition.}
\cref{tab:s2_eva} presents the quantitative results of the ablation study conducted on the \textit{flame salmon} scene, and more results are presented in Suppl.
The base model without Stage 2, and a set of randomly spawned 3DGs (Rnd. Spawn) in equivalent quantities to our spawn strategy, both fail to capture emerging objects.
The variant without our quantity control strategy (\textit{w/o} Quant. Ctrl.) manages to model emerging objects but requires a significantly larger number of additional 3DGs.
In contrast, our full model proficiently reconstructs emerging objects using a minimal addition of 3DGs.
The ablation study illustrated in \cref{fig:s2_abl} qualitatively showcases the effect of the Adaptive 3DG Addition strategy,
highlighting its ability to reconstruct the objects not present in the initial frame, such as coffee in a pot, a dog's tongue, and flames.

\textit{Real-time Rendering.}
Following 3DG-S~\cite{kerbl3Dgaussians}, we employ the SIBR framework~\cite{sibr2020} to measure the rendering speed.
Once all resources required are loaded onto the GPU, the additional overhead of our approach is primarily the time taken to query the NTC and transform the 3DGs.
As detailed in \cref{tab:rendering}, our method benefits from the efficiency of the multi-resolution hash encoding and the fully-fused MLP~\cite{tiny-cuda-nn}, which facilitate rapid NTC query.
Notably, the most time-consuming step is the SH Rotation.
However, our experiments indicate that the SH rotation has a minimal impact on the reconstruction quality,
which may be attributed to the 3DGs modeling view-dependent colors through alternative mechanisms (\eg, small 3DGs of varying colors surrounding the object) rather than SH coefficients.
Nonetheless, we maintain SH rotation for theoretical soundness.

%% file: 3DGStream/figs_tabs/NTC_eva.tex
\begin{figure}
    \centering
    \includegraphics[width=\linewidth]{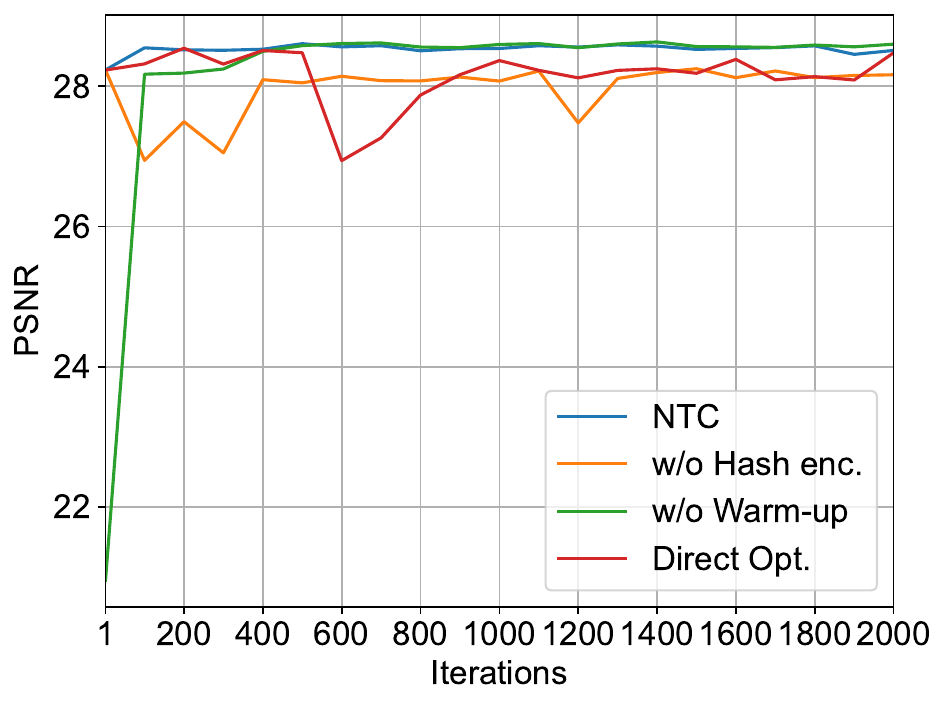}
    \caption{\textbf{Comparison of different approaches for modeling the transformation of 3DGs.}
    Conducted on the second frame of the \textit{flame salmon} video, utilizing identical initial 3DGs.}
\label{fig:NTC_eva}
\vspace{-3mm}
\end{figure}

%% file: 3DGStream/figs_tabs/s1_eva.tex
\begin{figure}
    \centering
    \includegraphics[width=\linewidth]{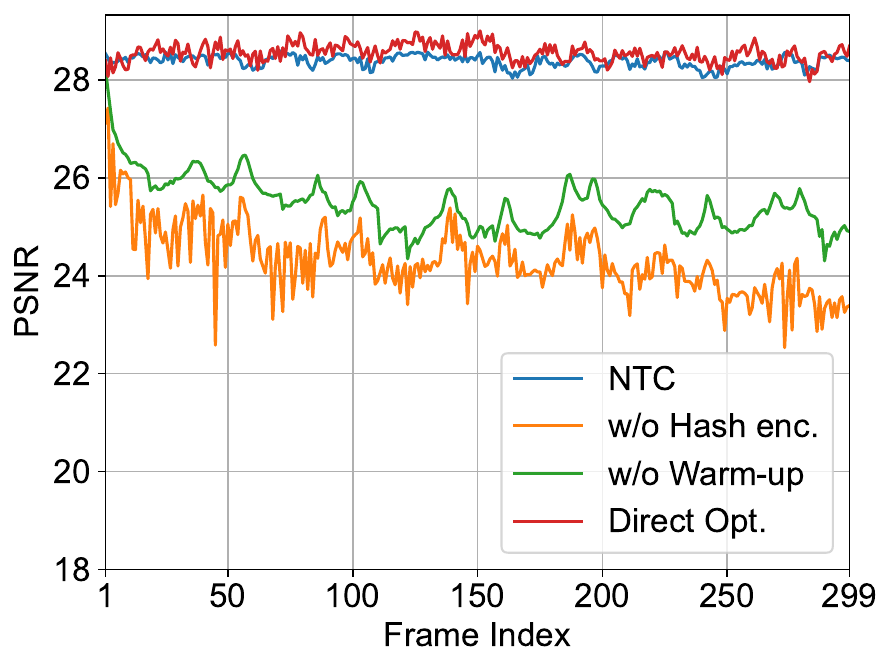}
    \caption{\textbf{Comparison of different approaches on the} \textit{flame salmon} \textbf{scene.}}
\label{fig:s1_eva}
\vspace{-3mm}
\end{figure}

%% file: 3DGStream/figs_tabs/s2_eva.tex
\begin{table}
    \centering
    \begin{tabular}{@{}l|cc@{}}
    \toprule
    Variant  & PSNR$\uparrow$ (dB)& \#Additional 3DGs$\downarrow$\\
    \midrule
    Baseline & 28.39 & \cellcolor{first}0      \\
    Rnd. Spawn  & 28.39 & 971.9     \\
    \textit{w/o} Quant. Ctrl.     & \cellcolor{first}28.43 & 8710.8       \\
    Full Model     & \cellcolor{second}28.42 & \cellcolor{second}477.7  \\
    \bottomrule
    \end{tabular}
    \caption{\textbf{Ablation study of the Adaptive 3DG Addition strategy on the}~\textit{flame salmon}~\textbf{scene}.
    The metrics are averaged over the whole sequence.}
    \label{tab:s2_eva}
    \vspace{-7.5mm}
\end{table}

%% file: 3DGStream/figs_tabs/s2_abl.tex
\begin{figure}
    \centering
    \begin{subfigure}{0.155\textwidth}
      \includegraphics[width=\textwidth]{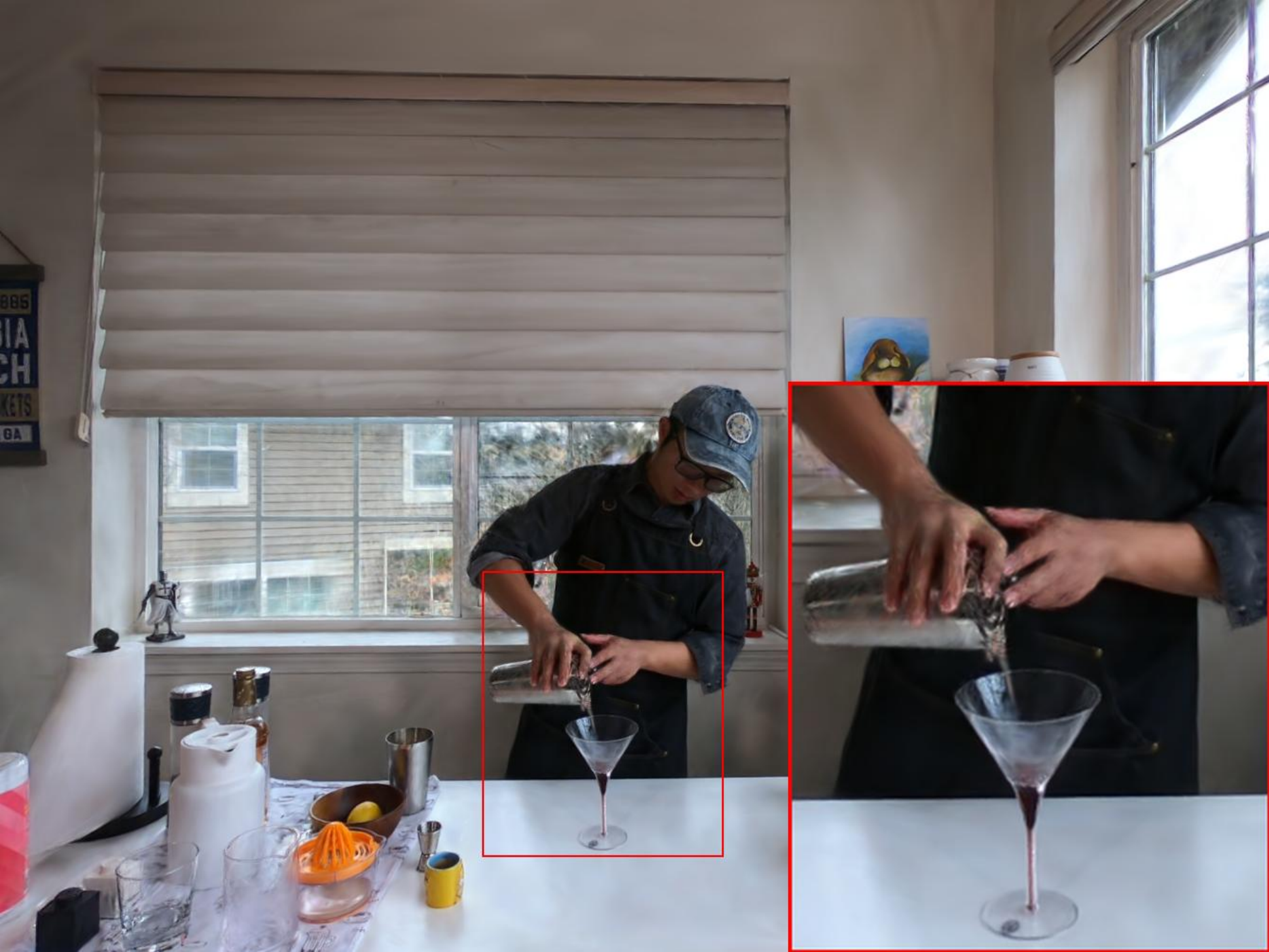}
    \end{subfigure}
    \hfill
    \begin{subfigure}{0.155\textwidth}
        \includegraphics[width=\textwidth]{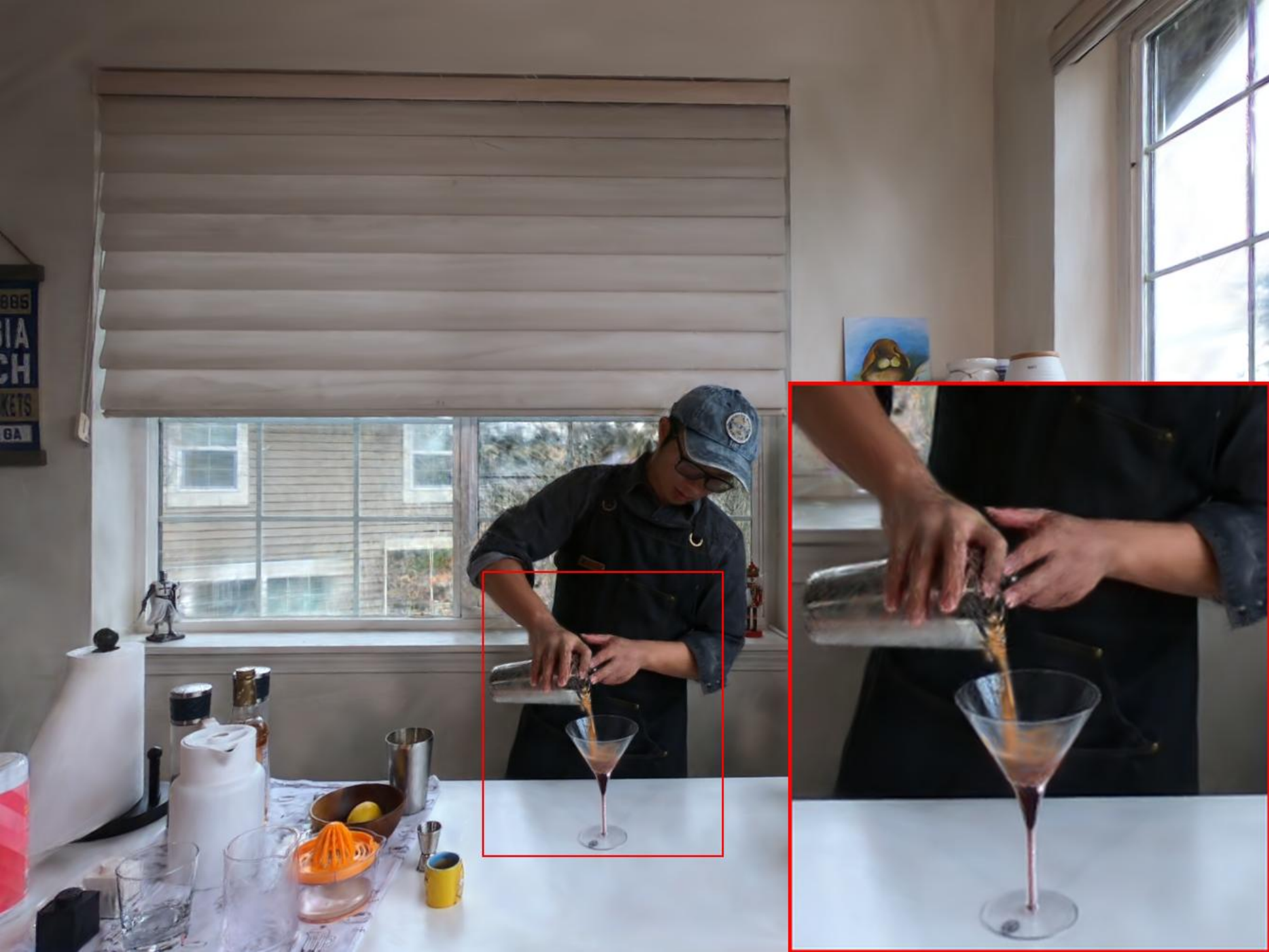}
    \end{subfigure}
    \hfill
    \begin{subfigure}{0.155\textwidth}
        \includegraphics[width=\textwidth]{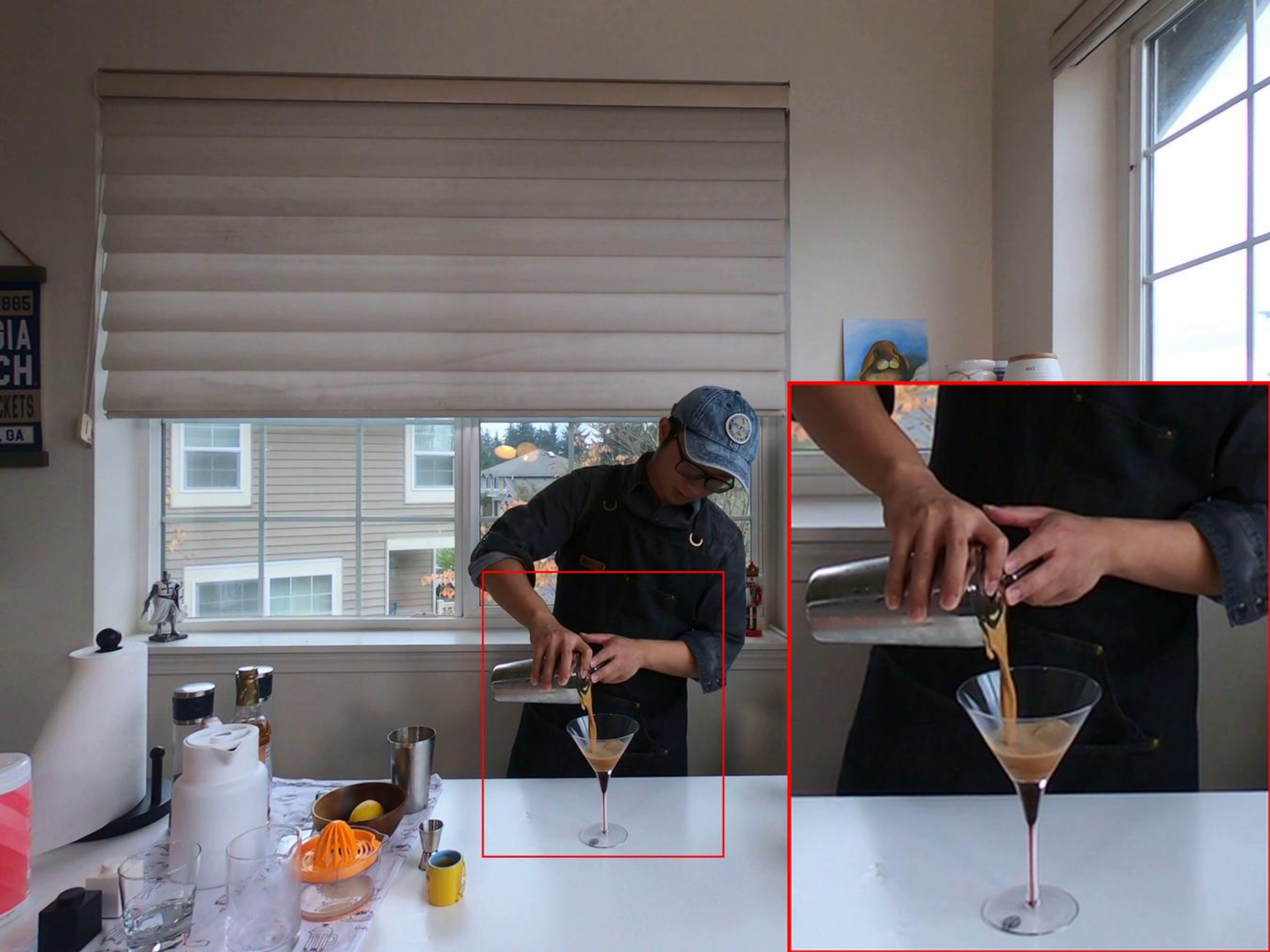}
    \end{subfigure}
    
    \vspace{0.05cm}
    
    \begin{subfigure}{0.155\textwidth}
      \includegraphics[width=\textwidth]{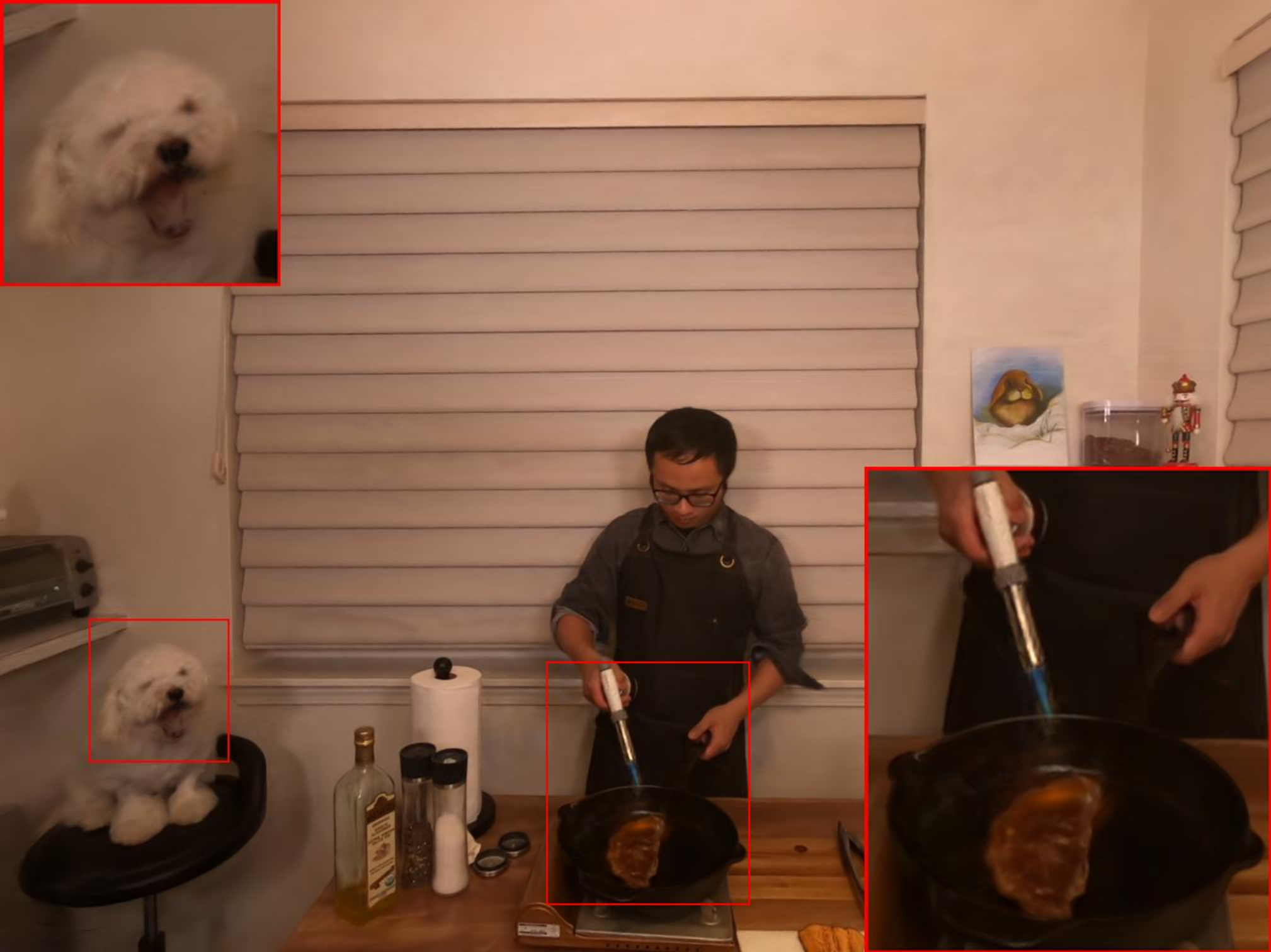}
        \caption*{(a) Result of Stage 1}
    \end{subfigure}
    \hfill
    \begin{subfigure}{0.155\textwidth}
        \includegraphics[width=\textwidth]{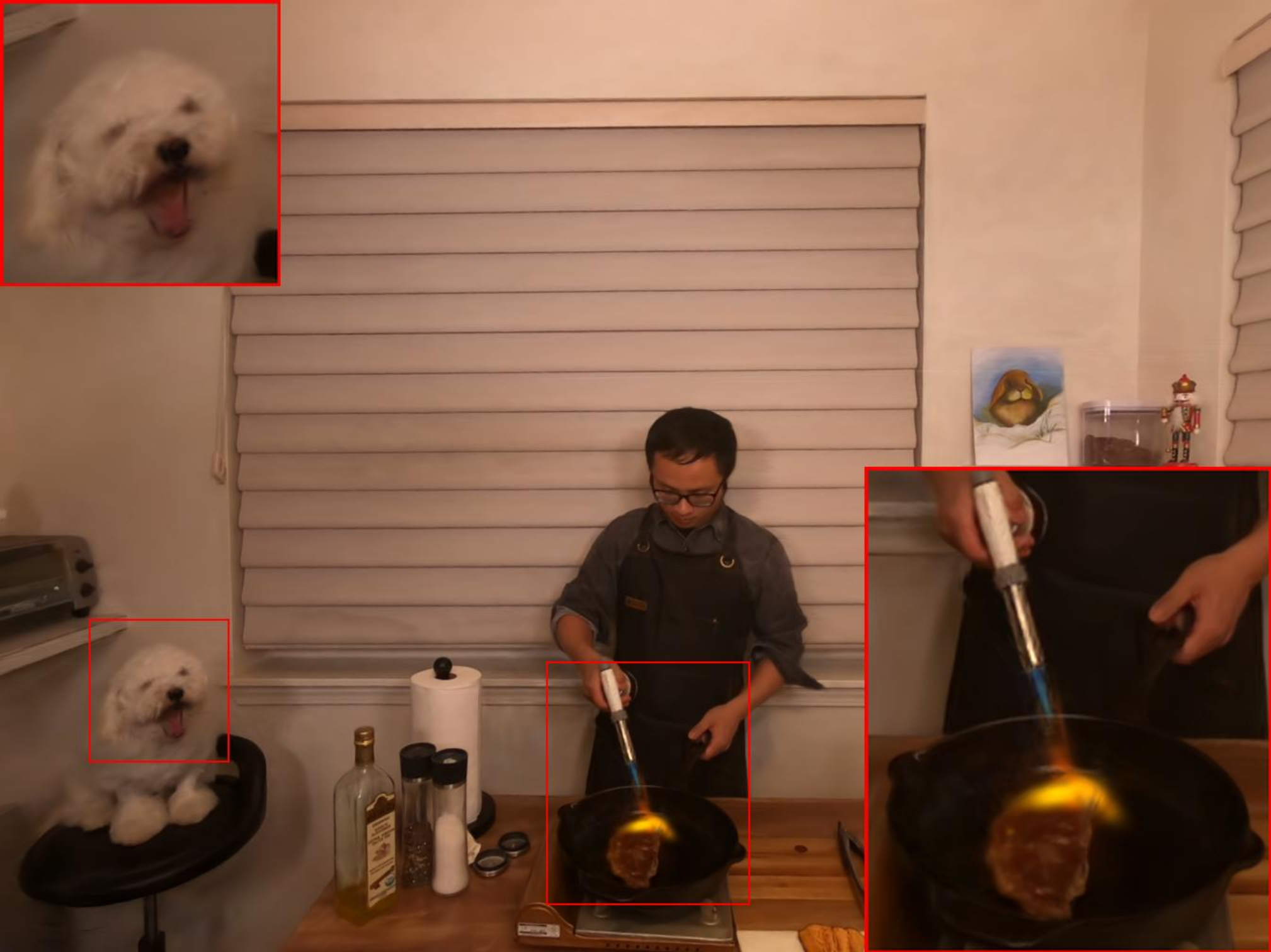}
        \caption*{(b) Result of Stage 2}
    \end{subfigure}
    \hfill
    \begin{subfigure}{0.155\textwidth}
        \includegraphics[width=\textwidth]{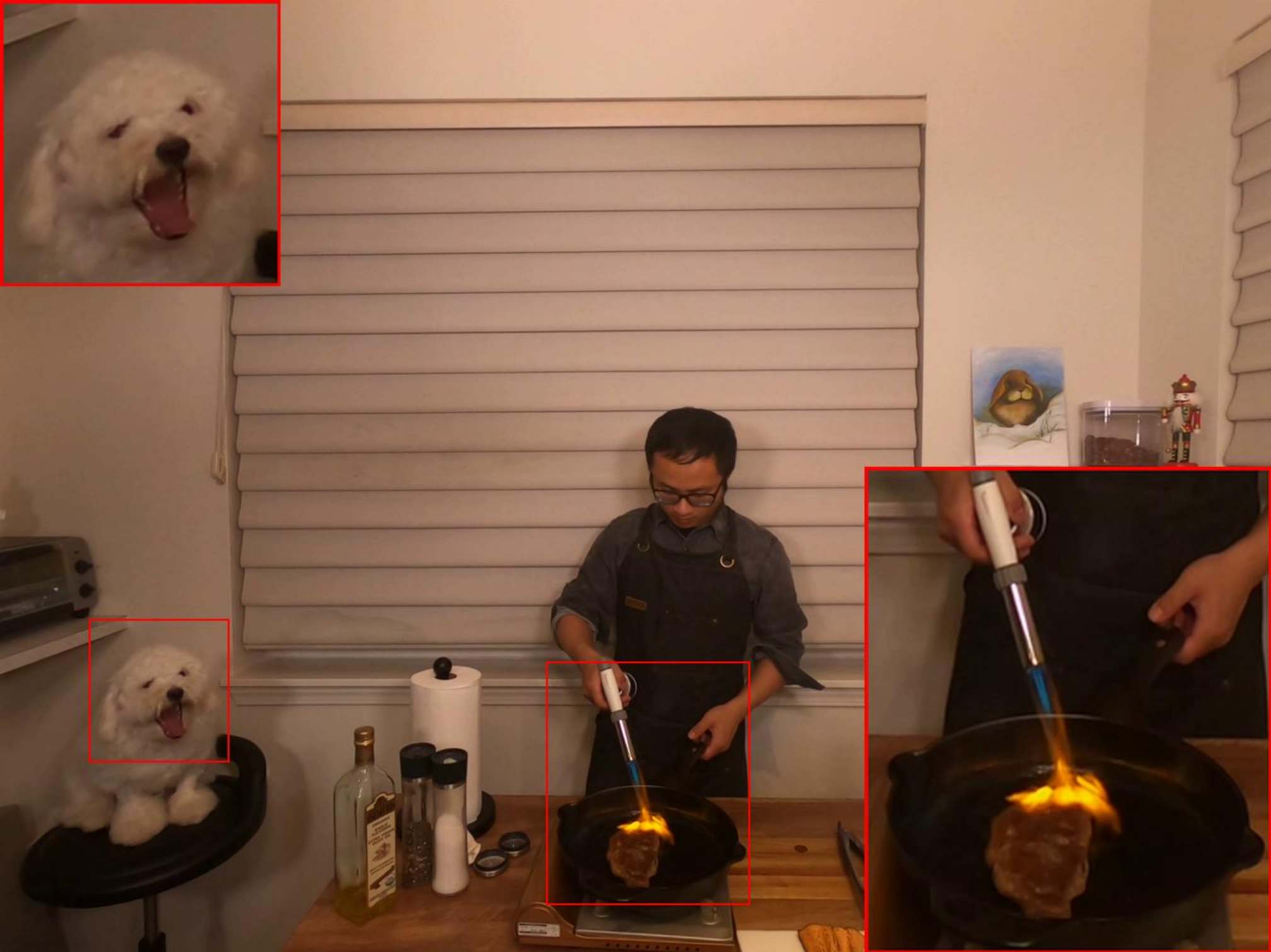}
        \caption*{(c) Ground Truth}
    \end{subfigure}
    \caption{\textbf{Quantitative results of the ablation study} conducted on the \textit{flame steak} scene and the \textit{coffee martini} scene.}
    \label{fig:s2_abl}
    \vspace{-5mm}
\end{figure}

%% file: 3DGStream/figs_tabs/rendering.tex
\begin{table}
    \centering
    \begin{tabular}{@{}l|cc@{}}
    \toprule
    Step & Overhead (ms) & FPS \\
    \midrule
    Render \textit{w/o} NTC & 2.56 & 390 \\
    \phantom{} + Query NTC & 0.62 & \\
    \phantom{} + Transformation & 0.02 & \\
    \phantom{} + SH Rotation & 1.46 & \\
    \midrule
    Total & 4.66 & 215 \\
    \bottomrule
    \end{tabular}
    \caption{\textbf{Rendering profilling} for the \textit{flame salmon} scene at megapixel resolution.
    Note that~\textit{flame salmon} is the most time-consuming to render of all scenes in our experiments.}
    \label{tab:rendering}
    \vspace{-5mm}
\end{table}

%% file: 3DGStream/6_limitations.tex
\section{Discussion}
\label{sec:discuss}
The quality of 3DG-S~\cite{kerbl3Dgaussians} on the initial frame is crucial to 3DGStream.
Therefore, we inherit the limitations of 3DG-S,
such as high dependence on the initial point cloud.
As illustrated in \cref{fig:s2_abl},
there are obvious artifacts beyond the windows, attributable to COLMAP's~\cite{schoenberger2016sfm} inability to reconstruct distant landscapes.
Hence, our method stands to benefit directly from future enhancements to 3DG-S.
Moreover, for efficient on-the-fly training,
we limit the number of training iterations,
which restricts modeling of drastic motion in Stage 1 and complex emerging objects in Stage 2.

%% file: 3DGStream/7_conclusion.tex
\section{Conclusion}
We propose 3DGStream, an novel method for efficient Free-Viewpoint Video streaming.
Based on 3DG-S~\cite{kerbl3Dgaussians}, we utilizes an effective Neural Transformation Cache to capture the motion of objects.
In addition, we propose an Adaptive 3DG Addition strategy to accurately model emerging objects in dynamic scenes.
The two-stage pipeline of 3DGStream enables the online reconstruction of dynamic scenes in video streams.
While ensuring photo-realistic image quality, 3DGStream achieves on-the-fly training ($\sim$10s per-frame) and real-time rendering ($\sim$200FPS) at megapixel resolution with moderate requisite storage.

\section{Acknowledgement}
This work was supported in part by Zhejiang Province Program (2022C01222, 2023C03199, 2023C03201),
the National Program of China (62172365, 2021YFF0900604, 19ZDA197),
Ningbo Science and Technology Plan Project (022Z167, 2023Z137),
and MOE Frontier Science Center for Brain Science \& Brain-Machine Integration (Zhejiang University).

%% file: 3DGStream/X_suppl.tex
\clearpage
\maketitlesupplementary
\section{Implementation Details}
\label{sec:appendix_impl}
We implement 3DGStream upon the codebase of 3D Gaussian Splatting (3DG-S)~\cite{kerbl3Dgaussians}
and use tiny-cuda-nn~\cite{tiny-cuda-nn} to implement Neural Transformation Cache (NTC).
All experiments were conducted on an NVIDIA RTX 3090 GPU.
In training of the initial frame, we let the densification of 3DG-S end at iteration 5000.
For the scenes in the N3DV dataset, we use the 3DGs of iteration 15000 as the initial 3DGs,
while for the scenes in the Meet Room dataset, we use the results of iteration 10000.
For convenience, we set the maximum degree of spherical harmonics (SH) to 1, and all other hyperparameters are consistent with the 3DG-S.

\textit{Training NTC}. We set the learning rate of NTC to 0.002.
For the scenes in the N3DV dataset~\cite{li2022dynerf}, the hash table size of the multi-resolution hash encoding is $2^{15}$,
the feature vector dimension is 4, and there are 16 resolution levels.
For the Meet Room dataset~\cite{li2022streaming}, the hash table length is $2^{14}$, with all other hyperparameters matching those specified for the N3DV dataset.
For all scenes, our fully-fused MLP comprises 2 hidden layers with 64 neurons each, employing ReLu as the activation function.
Given that the N3DV dataset and the Meet Room dataset both record indoor dynamic scenes,
and multi-resolution hash encoding requires normalized coordinates for input,
we create an axis-aligned bounding box that roughly encloses the house to normalize the 3D points
and discard any points outside the bounding box to prevent distant landscapes from influencing the training.

\textit{Training the additional 3DGs}. Compared to training on the initial frame, we increase the learning rate in the second stage for faster convergence.
Specifically, the learning rates for the mean, SH coefficient, opacity value, scaling vector,
and rotation quaternion of the 3DGs are set to 0.0024, 0.0375, 0.75, 0.075, and 0.015, respectively.
Note that these learning rates were not individually fine-tuned; instead, their proportions are following the default settings of 3DG-S.

\section{SH Rotation}
\label{sec:appendix_sh}
In order to preserve theoretical soundness, we also rotate the SH after transforming the 3DGs. The zeroth-degree SH does not require rotation; therefore, we only need to rotate the first-degree SH coefficients.

We utilize the projection function~\cite{sloan2008stupid} to project normal vectors onto the first-order SH. Given a rotation matrix $R$, we seek a matrix $M$ that can rotate the first-degree SH. Because rotating a vector before projecting it to SH produces the same outcome as projecting the vector first and then rotating the SH, we have the following relationship:
\begin{equation}
    MP(N)=P(RN),
\end{equation}
where $N$ is a normal vector.
For any three normal vectors $N_0, N_1$, and $N_2$ we denote $A=[P(N_0), P(N_1), P(N_2)]$. Consequently, we obtain:
\begin{equation}
    MA=[P(RN_0),P(RN_1),P(RN_2)].
\end{equation}
And hence:
\begin{equation}
    M=[P(RN_0),P(RN_1),P(RN_2)]A^{-1}
\end{equation}
For computational convenience, we choose $N_0=[1,0,0]^T, N_1=[0,1,0]^T$, and $N_2=[0,0,1]^T$.

\section{More Results}
\label{sec:appendix_res}
\input{3DGStream/figs_tabs/N3DV_comparisons}
\input{3DGStream/figs_tabs/N3DV_abl}
\input{3DGStream/figs_tabs/rendering_meetroom}
\subsection{Quantitative Results}
We provide a quantitative comparison of image quality,
measured by PSNR, across all scenes in the N3DV dataset in~\cref{tab:N3DV_Comparisons}.
Furthermore, we provide the quantitative result of the ablation study across all scenes in the N3DV dataset in~\cref{tab:N3DV_abl},
Additionally, we provide rendering profilling on the Meet Room Dataset in~\cref{tab:rendering_meetroom}.
\subsection{Qualitative Results}
We provide videos to show the free view synthesis results on various scenes from the N3DV dataset in \url{https://sjojok.github.io/3dgstream}.

\section{More Evaluations}
\label{sec:appendix_eval}
\input{3DGStream/figs_tabs/storage}
\input{3DGStream/figs_tabs/a3dg}
\input{3DGStream/figs_tabs/iter}
\subsection{Storage Requirements}
Except the initial 3DGs, we only need to store per-frame NTCs and per-frame additional 3DGs for each FVV frame, as detailed in~\cref{tab:rb}.
\subsection{Quantity of 3DGs}
In our experiments on the N3DV datasets, the quantity of initial 3DGs (\ie,~the transformed ones) is on the order of $10^5$, while the quantity of frame-specific additional 3DGs is on the order of $10^2$.
We show how the number of the frame-specific additional 3DGs changes as the frame number increases in~\cref{fig:rb}.
\subsection{Impact of Training Iterations}
In the main text, we discuss the trade-off between training efficiency and reconstruction quality, 
noting that limiting the number of training iterations enables efficient on-the-fly training at the expense of reduced quality.
To show this trade-off, we conduct experiments to evaluate the impact of training iterations, and show the quantitative results in~\cref{tab:iter}.
As shown in~\cref{tab:iter}, increasing training iterations in Stage 1 significantly enhances the reconstruction quality. 
However, an additional 100 iterations result in an increment of 3 seconds in the per-frame training duration. 
Incrementing training iterations in the second stage has a minimal impact on quality,
which can be attributed to the higher learning rate employed in this phase and the smaller number of additional 3DGs, facilitating rapid convergence.

%% file: 3DGStream/figs_tabs/N3DV_comparisons.tex
\begin{table}
    \centering
    \resizebox{\columnwidth}{!}{%
    \begin{tabular}{@{}l|c|c|c|c|c|c|c@{}}
      \toprule
      \multirow{2}{*}{Method} & {Coffee} & {Cook} & {Cut} & {Flame} & {Flame} & {Sear} & \multirow{2}{*}{Mean} \\
                              & {Martini} & {Spinach} & {Beef} & {Salmon} & {Steak} & {Steak} &  \\
      \midrule
      Plenoxels~\cite{fridovich2022plenoxels}$^\dagger$  & 27.65 & 31.73 & 32.01 & 28.68 & 32.24 & 32.33 & 30.77 \\
      I-NGP~\cite{muller2022instant}$^\dagger$      & 25.19 & {29.84} & 30.73 & 25.51 & {30.04} & 30.40 & {28.62} \\
      3DG-S~\cite{kerbl3Dgaussians}$^\dagger$       & 27.78 & \cellcolor{first}34.10 & \cellcolor{first}34.03 & 28.66 & \cellcolor{first}34.41 & \cellcolor{first}33.48 & \cellcolor{first}32.08 \\
      \midrule
      DyNeRF~\cite{li2022dynerf}     & {--} & {--} & {--} & 29.58 & {--} & {--} & 29.58 \\
      NeRFPlayer~\cite{song2022nerfplayer} & \cellcolor{first}31.53 & 30.58 & 29.35 & \cellcolor{first}31.65 & 31.93 & 29.13 & 30.69 \\
      HexPlane~\cite{Cao2022FWD}   & {--} & 32.04 & 32.55 & 29.47 & 32.08 & 32.39 & \cellcolor{second}31.70 \\
      K-Planes~\cite{kplanes_2023}   & \cellcolor{second}29.99 & \cellcolor{third}32.60 & 31.82 & \cellcolor{second}30.44 & \cellcolor{third}32.38 & 32.52 & 31.63 \\
      HyperReel~\cite{attal2023hyperreel}  & 28.37 & 32.30 & \cellcolor{third}32.92 & 28.26 & 32.20 & \cellcolor{third}32.57 & 31.10 \\
      MixVoxels~\cite{wang2023mixed} & \cellcolor{third}29.36 & 31.61 & 31.30 & \cellcolor{third}29.92 & 31.43 & 31.21 & 30.80 \\
      \midrule
      StreamRF~\cite{li2022streaming}$^\dagger$   & 27.84 & 31.59 & 31.81 & 28.26 & 32.24 & 32.36 & 28.26 \\
      Ours       & 27.75 & \cellcolor{second}33.31 & \cellcolor{second}33.21 & 28.42 & \cellcolor{second}34.30 & \cellcolor{second}33.01 & \cellcolor{third}31.67 \\
      \bottomrule
    \end{tabular}
    }
    \caption{\textbf{Quantitative comparison} of PSNR values across all scenes in the N3DV dataset, with the metric for each scene calculated as the average over 300 frames.
    $^\dagger$Obtained in our own experiments with the official codes.
    }
    \label{tab:N3DV_Comparisons}
\end{table}

%% file: 3DGStream/figs_tabs/N3DV_abl.tex
\begin{table}
    \centering
    \resizebox{\columnwidth}{!}{%
    \begin{tabular}{@{}l|c|c|c|c|c|c|c@{}}
      \toprule
      \multirow{2}{*}{Method} & {Coffee} & {Cook} & {Cut} & {Flame} & {Flame} & {Sear} & \multirow{2}{*}{Mean} \\
                              & {Martini} & {Spinach} & {Beef} & {Salmon} & {Steak} & {Steak} &  \\
      \midrule
      Baseline   & 27.68 & 33.19 & 33.10 & 28.39 & 33.54 & 32.79 & 31.45 \\
      Full Model   & 27.75 & 33.31 & 33.21 & 28.42 & 34.30 & 33.01 & 31.67 \\
      \bottomrule
    \end{tabular}
    }
    \caption{\textbf{Ablation Study} of the Adaptive 3DG Addition strategy across all scenes in the N3DV dataset, with the metric for each scene calculated as the average over 300 frames.
    We take PSNR to measure the image quality.}
    \label{tab:N3DV_abl}
\end{table}

%% file: 3DGStream/figs_tabs/rendering_meetroom.tex
\begin{table}
    \centering
    \begin{tabular}{@{}l|cc@{}}
    \toprule
    Step & Overhead (ms) & FPS \\
    \midrule
    Render \textit{w/o} NTC            & 1.75           & 571          \\
    + Query NTC                        & +0.46          &              \\
    + Transformation                   & +0.02          &              \\
    + SH Rotation                      & +1.24          &              \\
    \midrule
    Total                              & 3.47           & 288 \\
    \bottomrule
    \end{tabular}
    \caption{\textbf{Rendering profilling} on the Meet Room dataset.}
    \label{tab:rendering_meetroom}
\end{table}
    

%% file: 3DGStream/figs_tabs/storage.tex
\begin{table}
    \centering
    \resizebox{\linewidth}{!}{
    \begin{tabular}{l|cc|c}
    \toprule
    Dataset  & NTC (KB) & New 3DGs (KB) & Total (KB)\\
    \midrule
    N3DV &7781.5 & 49.1    & 7830.6\\
    MeetRoom &3941.5 & 195.3    & 4136.8\\
    \bottomrule
    \end{tabular}
    }
    \caption{\textbf{Detailed ``Storage'' entry of our method} in Tabs. 1 and 2.}
    \label{tab:rb}
    \vspace{-4mm}
\end{table}

%% file: 3DGStream/figs_tabs/a3dg.tex
\begin{figure}[t]
    \centering
    \includegraphics[width=\linewidth]{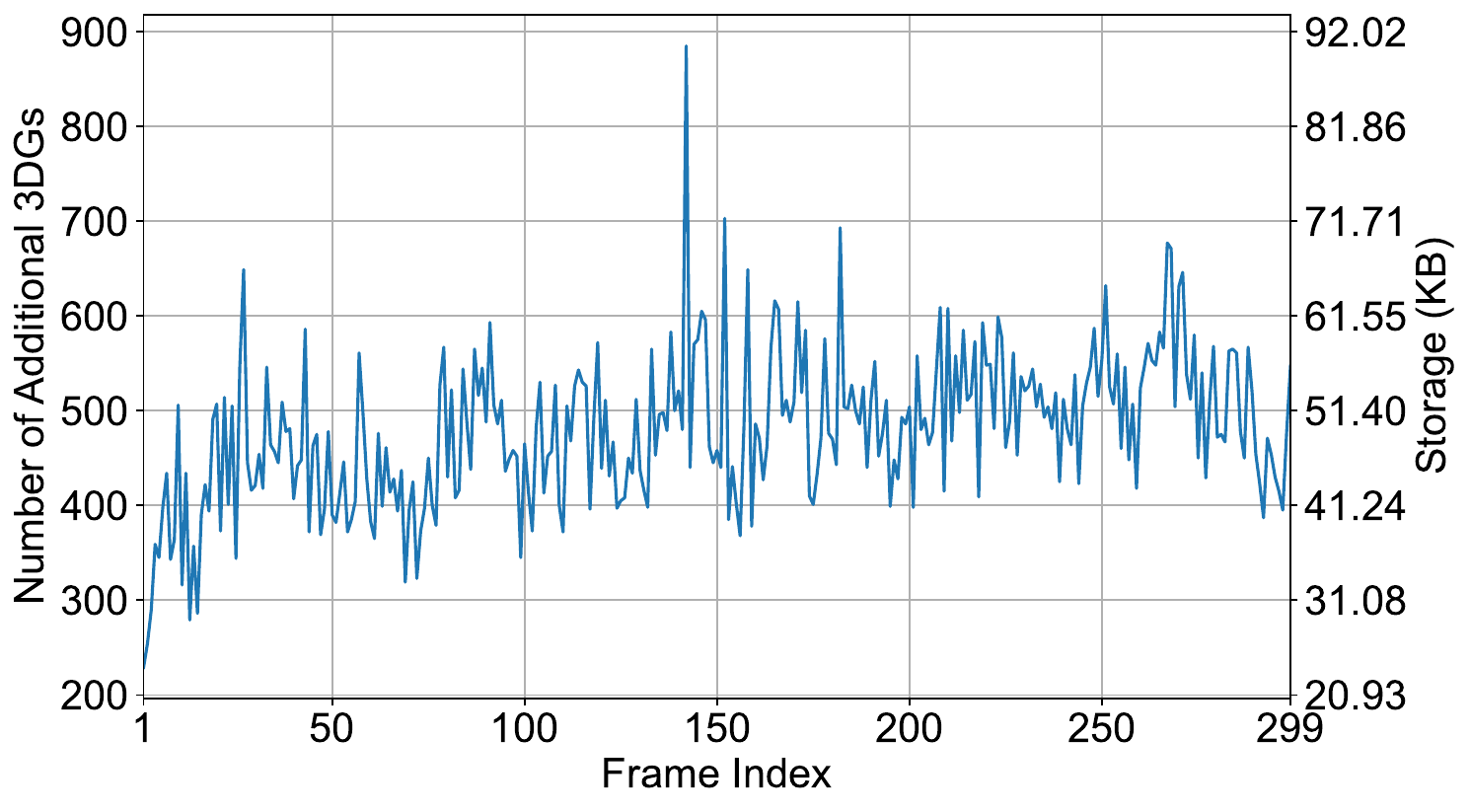}
     \caption{\textbf{Number of additional 3DGs and corresponding storage requirement} of each frame on the \textit{flame salmon} scene.}
     \label{fig:rb}
     \vspace{-6.5mm}
\end{figure}

%% file: 3DGStream/figs_tabs/iter.tex
\begin{table}[]
    \centering
    \begin{tabular}{@{}l|SSSSSS@{}}
    \toprule
    {\multirow{2}{*}{Scene}} & \multicolumn{2}{c}{Stage 1} & \multicolumn{2}{c}{Stage 2} \\
    \cmidrule(lr){2-3} \cmidrule(lr){4-5}
    & {150} & {250} & {100} & {200} \\
    \midrule
    Flame Salmon       & 28.39 & 28.44 & 28.46 & 28.46 \\
    Flame Steak      & 33.54 & 33.81 & 34.44 & 34.46 \\
    Sear Steak       & 32.79 & 33.02 & 33.18 & 33.19 \\
    Cook Spinach     & 33.19 & 33.50 & 33.56 & 33.57 \\
    Cut Roasted Beef & 33.10 & 33.39 & 33.44 & 33.44 \\
    Coffee Martini   & 27.68 & 27.77 & 27.83 & 27.83 \\
    \bottomrule
    \end{tabular}
    \caption{\textbf{Evaluation on the impact of training iterations} conducted on the N3DV dataset.
    The result of Stage 2 is is obtained after 250 iterations of optimization at Stage 1.
    We take PSNR to measure the image quality.}
    \label{tab:iter}
\end{table}

%% file: main.bbl
\begin{thebibliography}{77}
\providecommand{\natexlab}[1]{#1}
\providecommand{\url}[1]{\texttt{#1}}
\expandafter\ifx\csname urlstyle\endcsname\relax
  \providecommand{\doi}[1]{doi: #1}\else
  \providecommand{\doi}{doi: \begingroup \urlstyle{rm}\Url}\fi

\bibitem[Attal et~al.(2023)Attal, Huang, Richardt, Zollhoefer, Kopf, O’Toole,
  and Kim]{attal2023hyperreel}
Benjamin Attal, Jia-Bin Huang, Christian Richardt, Michael Zollhoefer, Johannes
  Kopf, Matthew O’Toole, and Changil Kim.
\newblock Hyperreel: High-fidelity 6-dof video with ray-conditioned sampling.
\newblock In \emph{Proceedings of the IEEE/CVF Conference on Computer Vision
  and Pattern Recognition}, pages 16610--16620, 2023.

\bibitem[Barron et~al.(2021)Barron, Mildenhall, Tancik, Hedman, Martin-Brualla,
  and Srinivasan]{barron2021mip}
Jonathan~T Barron, Ben Mildenhall, Matthew Tancik, Peter Hedman, Ricardo
  Martin-Brualla, and Pratul~P Srinivasan.
\newblock Mip-nerf: A multiscale representation for anti-aliasing neural
  radiance fields.
\newblock In \emph{Proceedings of the IEEE/CVF International Conference on
  Computer Vision}, pages 5855--5864, 2021.

\bibitem[Barron et~al.(2022)Barron, Mildenhall, Verbin, Srinivasan, and
  Hedman]{barron2022mip}
Jonathan~T Barron, Ben Mildenhall, Dor Verbin, Pratul~P Srinivasan, and Peter
  Hedman.
\newblock Mip-nerf 360: Unbounded anti-aliased neural radiance fields.
\newblock In \emph{Proceedings of the IEEE/CVF Conference on Computer Vision
  and Pattern Recognition}, pages 5470--5479, 2022.

\bibitem[Barron et~al.(2023)Barron, Mildenhall, Verbin, Srinivasan, and
  Hedman]{barron2023zipnerf}
Jonathan~T. Barron, Ben Mildenhall, Dor Verbin, Pratul~P. Srinivasan, and Peter
  Hedman.
\newblock Zip-nerf: Anti-aliased grid-based neural radiance fields.
\newblock \emph{ICCV}, 2023.

\bibitem[Black and Anandan(1996)]{black1996robust}
Michael~J Black and Paul Anandan.
\newblock The robust estimation of multiple motions: Parametric and
  piecewise-smooth flow fields.
\newblock \emph{Computer vision and image understanding}, 63\penalty0
  (1):\penalty0 75--104, 1996.

\bibitem[Bonopera et~al.(2020)Bonopera, Esnault, Prakash, Rodriguez, Thonat,
  Benadel, Chaurasia, Philip, and Drettakis]{sibr2020}
Sebastien Bonopera, Jerome Esnault, Siddhant Prakash, Simon Rodriguez, Theo
  Thonat, Mehdi Benadel, Gaurav Chaurasia, Julien Philip, and George Drettakis.
\newblock sibr: A system for image based rendering, 2020.

\bibitem[Broxton et~al.(2020)Broxton, Flynn, Overbeck, Erickson, Hedman,
  Duvall, Dourgarian, Busch, Whalen, and Debevec]{broxton2020immersive}
Michael Broxton, John Flynn, Ryan Overbeck, Daniel Erickson, Peter Hedman,
  Matthew Duvall, Jason Dourgarian, Jay Busch, Matt Whalen, and Paul Debevec.
\newblock Immersive light field video with a layered mesh representation.
\newblock \emph{ACM Transactions on Graphics (TOG)}, 39\penalty0 (4):\penalty0
  86--1, 2020.

\bibitem[Buehler et~al.(2001)Buehler, Bosse, McMillan, Gortler, and
  Cohen]{buehler2001unstructured}
Chris Buehler, Michael Bosse, Leonard McMillan, Steven Gortler, and Michael
  Cohen.
\newblock Unstructured lumigraph rendering.
\newblock In \emph{SIGGRAPH}, pages 425--432, 2001.

\bibitem[Cao and Johnson(2023)]{Cao2022FWD}
Ang Cao and Justin Johnson.
\newblock Hexplane: A fast representation for dynamic scenes.
\newblock \emph{CVPR}, 2023.

\bibitem[Chai et~al.(2000)Chai, Tong, Chan, and Shum]{chai2000plenoptic}
Jin-Xiang Chai, Xin Tong, Shing-Chow Chan, and Heung-Yeung Shum.
\newblock Plenoptic sampling.
\newblock In \emph{Proceedings of the 27th annual conference on Computer
  graphics and interactive techniques}, pages 307--318, 2000.

\bibitem[Chen et~al.(2021)Chen, Xu, Zhao, Zhang, Xiang, Yu, and
  Su]{chen2021mvsnerf}
Anpei Chen, Zexiang Xu, Fuqiang Zhao, Xiaoshuai Zhang, Fanbo Xiang, Jingyi Yu,
  and Hao Su.
\newblock Mvsnerf: Fast generalizable radiance field reconstruction from
  multi-view stereo.
\newblock In \emph{Proceedings of the IEEE/CVF International Conference on
  Computer Vision (ICCV)}, pages 14124--14133, 2021.

\bibitem[Chen et~al.(2022)Chen, Xu, Geiger, Yu, and Su]{chen2022tensorf}
Anpei Chen, Zexiang Xu, Andreas Geiger, Jingyi Yu, and Hao Su.
\newblock Tensorf: Tensorial radiance fields.
\newblock In \emph{European Conference on Computer Vision (ECCV)}, 2022.

\bibitem[Chen et~al.(2023{\natexlab{a}})Chen, Xu, Wei, Tang, Su, and
  Geiger]{Chen2023TOG}
Anpei Chen, Zexiang Xu, Xinyue Wei, Siyu Tang, Hao Su, and Andreas Geiger.
\newblock Dictionary fields: Learning a neural basis decomposition.
\newblock \emph{ACM Trans. Graph.}, 2023{\natexlab{a}}.

\bibitem[Chen et~al.(2023{\natexlab{b}})Chen, Funkhouser, Hedman, and
  Tagliasacchi]{chen2022mobilenerf}
Zhiqin Chen, Thomas Funkhouser, Peter Hedman, and Andrea Tagliasacchi.
\newblock Mobilenerf: Exploiting the polygon rasterization pipeline for
  efficient neural field rendering on mobile architectures.
\newblock In \emph{The Conference on Computer Vision and Pattern Recognition
  (CVPR)}, 2023{\natexlab{b}}.

\bibitem[Collet et~al.(2015)Collet, Chuang, Sweeney, Gillett, Evseev,
  Calabrese, Hoppe, Kirk, and Sullivan]{collet2015high}
Alvaro Collet, Ming Chuang, Pat Sweeney, Don Gillett, Dennis Evseev, David
  Calabrese, Hugues Hoppe, Adam Kirk, and Steve Sullivan.
\newblock High-quality streamable free-viewpoint video.
\newblock \emph{ACM Transactions on Graphics (TOG)}, 34\penalty0 (4):\penalty0
  69, 2015.

\bibitem[Davis et~al.(2012)Davis, Levoy, and Durand]{davis2012unstructured}
Abe Davis, Marc Levoy, and Fredo Durand.
\newblock Unstructured light fields.
\newblock \emph{Comput. Graph. Forum}, 31\penalty0 (2pt1):\penalty0 305–314,
  2012.

\bibitem[Dou et~al.(2017)Dou, Davidson, Fanello, Khamis, Kowdle, Rhemann,
  Tankovich, and Izadi]{motion2fusion}
Mingsong Dou, Philip Davidson, Sean~Ryan Fanello, Sameh Khamis, Adarsh Kowdle,
  Christoph Rhemann, Vladimir Tankovich, and Shahram Izadi.
\newblock Motion2fusion: Real-time volumetric performance capture.
\newblock \emph{ACM Trans. Graph.}, 36\penalty0 (6):\penalty0 246:1--246:16,
  2017.

\bibitem[Duan et~al.(2024)Duan, Wei, Dai, He, Chen, and Chen]{duan20244d}
Yuanxing Duan, Fangyin Wei, Qiyu Dai, Yuhang He, Wenzheng Chen, and Baoquan
  Chen.
\newblock 4d gaussian splatting: Towards efficient novel view synthesis for
  dynamic scenes.
\newblock \emph{arXiv preprint arXiv:2402.03307}, 2024.

\bibitem[Fang et~al.(2022)Fang, Yi, Wang, Xie, Zhang, Liu, Nie\ss{}ner, and
  Tian]{TiNeuVox}
Jiemin Fang, Taoran Yi, Xinggang Wang, Lingxi Xie, Xiaopeng Zhang, Wenyu Liu,
  Matthias Nie\ss{}ner, and Qi Tian.
\newblock Fast dynamic radiance fields with time-aware neural voxels.
\newblock In \emph{SIGGRAPH Asia 2022 Conference Papers}, 2022.

\bibitem[Fridovich-Keil et~al.(2022)Fridovich-Keil, Yu, Tancik, Chen, Recht,
  and Kanazawa]{fridovich2022plenoxels}
Sara Fridovich-Keil, Alex Yu, Matthew Tancik, Qinhong Chen, Benjamin Recht, and
  Angjoo Kanazawa.
\newblock Plenoxels: Radiance fields without neural networks.
\newblock In \emph{Proceedings of the IEEE/CVF Conference on Computer Vision
  and Pattern Recognition}, pages 5501--5510, 2022.

\bibitem[Garbin et~al.(2021)Garbin, Kowalski, Johnson, Shotton, and
  Valentin]{garbin2021fastnerf}
Stephan~J. Garbin, Marek Kowalski, Matthew Johnson, Jamie Shotton, and Julien
  Valentin.
\newblock Fastnerf: High-fidelity neural rendering at 200fps.
\newblock In \emph{Proceedings of the IEEE/CVF International Conference on
  Computer Vision (ICCV)}, pages 14346--14355, 2021.

\bibitem[Gortler et~al.(1996)Gortler, Grzeszczuk, Szeliski, and
  Cohen]{gortler1996lumigraph}
Steven~J. Gortler, Radek Grzeszczuk, Richard Szeliski, and Michael~F. Cohen.
\newblock The lumigraph.
\newblock In \emph{Proceedings of the 23rd Annual Conference on Computer
  Graphics and Interactive Techniques}, page 43–54, New York, NY, USA, 1996.
  Association for Computing Machinery.

\bibitem[Hedman et~al.(2021)Hedman, Srinivasan, Mildenhall, Barron, and
  Debevec]{hedman2021snerg}
Peter Hedman, Pratul~P. Srinivasan, Ben Mildenhall, Jonathan~T. Barron, and
  Paul Debevec.
\newblock Baking neural radiance fields for real-time view synthesis.
\newblock In \emph{2021 IEEE/CVF International Conference on Computer Vision
  (ICCV)}, pages 5855--5864, 2021.

\bibitem[Horn and Schunck(1981)]{horn1981determining}
Berthold~KP Horn and Brian~G Schunck.
\newblock Determining optical flow.
\newblock \emph{Artificial intelligence}, 17\penalty0 (1-3):\penalty0 185--203,
  1981.

\bibitem[Hu et~al.(2023)Hu, Wang, Ma, Yang, Gao, Liu, and Ma]{hu2023Tri-MipRF}
Wenbo Hu, Yuling Wang, Lin Ma, Bangbang Yang, Lin Gao, Xiao Liu, and Yuewen Ma.
\newblock Tri-miprf: Tri-mip representation for efficient anti-aliasing neural
  radiance fields.
\newblock In \emph{ICCV}, 2023.

\bibitem[Kerbl et~al.(2023)Kerbl, Kopanas, Leimk{\"u}hler, and
  Drettakis]{kerbl3Dgaussians}
Bernhard Kerbl, Georgios Kopanas, Thomas Leimk{\"u}hler, and George Drettakis.
\newblock 3d gaussian splatting for real-time radiance field rendering.
\newblock \emph{ACM Transactions on Graphics}, 42\penalty0 (4), 2023.

\bibitem[Kirschstein et~al.(2023)Kirschstein, Qian, Giebenhain, Walter, and
  Nie{\ss}ner]{kirschstein2023nersemble}
Tobias Kirschstein, Shenhan Qian, Simon Giebenhain, Tim Walter, and Matthias
  Nie{\ss}ner.
\newblock Nersemble: Multi-view radiance field reconstruction of human heads.
\newblock \emph{arXiv preprint arXiv:2305.03027}, 2023.

\bibitem[Levoy and Hanrahan(1996)]{levoy1996light}
Marc Levoy and Pat Hanrahan.
\newblock Light field rendering.
\newblock In \emph{SIGGRAPH}, pages 31--42, 1996.

\bibitem[Li et~al.(2022{\natexlab{a}})Li, Shen, Wang, Shen, and
  Tan]{li2022streaming}
Lingzhi Li, Zhen Shen, Zhongshu Wang, Li Shen, and Ping Tan.
\newblock Streaming radiance fields for 3d video synthesis.
\newblock In \emph{NeurIPS}, 2022{\natexlab{a}}.

\bibitem[Li et~al.(2022{\natexlab{b}})Li, Tanke, Vo, Zollh{\"o}fer, Gall,
  Kanazawa, and Lassner]{li2022tava}
Ruilong Li, Julian Tanke, Minh Vo, Michael Zollh{\"o}fer, J{\"u}rgen Gall,
  Angjoo Kanazawa, and Christoph Lassner.
\newblock Tava: Template-free animatable volumetric actors.
\newblock In \emph{European Conference on Computer Vision}, pages 419--436.
  Springer, 2022{\natexlab{b}}.

\bibitem[Li et~al.(2022{\natexlab{c}})Li, Slavcheva, Zollhoefer, Green,
  Lassner, Kim, Schmidt, Lovegrove, Goesele, Newcombe, et~al.]{li2022dynerf}
Tianye Li, Mira Slavcheva, Michael Zollhoefer, Simon Green, Christoph Lassner,
  Changil Kim, Tanner Schmidt, Steven Lovegrove, Michael Goesele, Richard
  Newcombe, et~al.
\newblock Neural 3d video synthesis from multi-view video.
\newblock In \emph{Proceedings of the IEEE/CVF Conference on Computer Vision
  and Pattern Recognition}, pages 5521--5531, 2022{\natexlab{c}}.

\bibitem[Li et~al.(2021)Li, Niklaus, Snavely, and Wang]{li2020neural}
Zhengqi Li, Simon Niklaus, Noah Snavely, and Oliver Wang.
\newblock Neural scene flow fields for space-time view synthesis of dynamic
  scenes.
\newblock In \emph{Proceedings of the IEEE/CVF Conference on Computer Vision
  and Pattern Recognition (CVPR)}, pages 6498--6508, 2021.

\bibitem[Li et~al.(2023)Li, Wang, Cole, Tucker, and Snavely]{li2023dynibar}
Zhengqi Li, Qianqian Wang, Forrester Cole, Richard Tucker, and Noah Snavely.
\newblock Dynibar: Neural dynamic image-based rendering.
\newblock In \emph{Proceedings of the IEEE/CVF Conference on Computer Vision
  and Pattern Recognition (CVPR)}, 2023.

\bibitem[Luiten et~al.(2024)Luiten, Kopanas, Leibe, and
  Ramanan]{luiten2023dynamic}
Jonathon Luiten, Georgios Kopanas, Bastian Leibe, and Deva Ramanan.
\newblock Dynamic 3d gaussians: Tracking by persistent dynamic view synthesis.
\newblock In \emph{3DV}, 2024.

\bibitem[Martin-Brualla et~al.(2021)Martin-Brualla, Radwan, Sajjadi, Barron,
  Dosovitskiy, and Duckworth]{martinbrualla2020nerfw}
Ricardo Martin-Brualla, Noha Radwan, Mehdi S.~M. Sajjadi, Jonathan~T. Barron,
  Alexey Dosovitskiy, and Daniel Duckworth.
\newblock {NeRF in the Wild: Neural Radiance Fields for Unconstrained Photo
  Collections}.
\newblock In \emph{CVPR}, 2021.

\bibitem[Mildenhall et~al.(2020)Mildenhall, Srinivasan, Tancik, Barron,
  Ramamoorthi, and Ng]{mildenhall2020nerf}
Ben Mildenhall, Pratul~P Srinivasan, Matthew Tancik, Jonathan~T Barron, Ravi
  Ramamoorthi, and Ren Ng.
\newblock Nerf: Representing scenes as neural radiance fields for view
  synthesis.
\newblock In \emph{European conference on computer vision}, pages 405--421.
  Springer, 2020.

\bibitem[Mildenhall et~al.(2022)Mildenhall, Hedman, Martin-Brualla, Srinivasan,
  and Barron]{mildenhall2022nerf}
Ben Mildenhall, Peter Hedman, Ricardo Martin-Brualla, Pratul~P Srinivasan, and
  Jonathan~T Barron.
\newblock Nerf in the dark: High dynamic range view synthesis from noisy raw
  images.
\newblock In \emph{Proceedings of the IEEE/CVF Conference on Computer Vision
  and Pattern Recognition}, pages 16190--16199, 2022.

\bibitem[M\"uller(2021)]{tiny-cuda-nn}
Thomas M\"uller.
\newblock {tiny-cuda-nn}, 2021.

\bibitem[M{\"u}ller et~al.(2021)M{\"u}ller, Rousselle, Nov{\'a}k, and
  Keller]{muller2021real}
Thomas M{\"u}ller, Fabrice Rousselle, Jan Nov{\'a}k, and Alexander Keller.
\newblock Real-time neural radiance caching for path tracing.
\newblock \emph{ACM Transactions on Graphics (TOG)}, 40\penalty0 (4):\penalty0
  1--16, 2021.

\bibitem[M\"uller et~al.(2022)M\"uller, Evans, Schied, and
  Keller]{muller2022instant}
Thomas M\"uller, Alex Evans, Christoph Schied, and Alexander Keller.
\newblock Instant neural graphics primitives with a multiresolution hash
  encoding.
\newblock \emph{ACM Trans. Graph.}, 41\penalty0 (4):\penalty0 102:1--102:15,
  2022.

\bibitem[Niemeyer et~al.(2022)Niemeyer, Barron, Mildenhall, Sajjadi, Geiger,
  and Radwan]{niemeyer2022regnerf}
Michael Niemeyer, Jonathan~T Barron, Ben Mildenhall, Mehdi~SM Sajjadi, Andreas
  Geiger, and Noha Radwan.
\newblock Regnerf: Regularizing neural radiance fields for view synthesis from
  sparse inputs.
\newblock In \emph{Proceedings of the IEEE/CVF Conference on Computer Vision
  and Pattern Recognition}, pages 5480--5490, 2022.

\bibitem[Park et~al.(2021{\natexlab{a}})Park, Sinha, Barron, Bouaziz, Goldman,
  Seitz, and Martin-Brualla]{park2020deformable}
Keunhong Park, Utkarsh Sinha, Jonathan~T. Barron, Sofien Bouaziz, Dan~B
  Goldman, Steven~M. Seitz, and Ricardo Martin-Brualla.
\newblock Nerfies: Deformable neural radiance fields.
\newblock In \emph{Proceedings of the IEEE/CVF International Conference on
  Computer Vision (ICCV)}, pages 5865--5874, 2021{\natexlab{a}}.

\bibitem[Park et~al.(2021{\natexlab{b}})Park, Sinha, Barron, Bouaziz, Goldman,
  Seitz, and Martin-Brualla]{park2021nerfies}
Keunhong Park, Utkarsh Sinha, Jonathan~T Barron, Sofien Bouaziz, Dan~B Goldman,
  Steven~M Seitz, and Ricardo Martin-Brualla.
\newblock Nerfies: Deformable neural radiance fields.
\newblock In \emph{Proceedings of the IEEE/CVF International Conference on
  Computer Vision}, pages 5865--5874, 2021{\natexlab{b}}.

\bibitem[Park et~al.(2021{\natexlab{c}})Park, Sinha, Hedman, Barron, Bouaziz,
  Goldman, Martin-Brualla, and Seitz]{park2021hypernerf}
Keunhong Park, Utkarsh Sinha, Peter Hedman, Jonathan~T. Barron, Sofien Bouaziz,
  Dan~B Goldman, Ricardo Martin-Brualla, and Steven~M. Seitz.
\newblock Hypernerf: A higher-dimensional representation for topologically
  varying neural radiance fields.
\newblock \emph{ACM Trans. Graph.}, 40\penalty0 (6), 2021{\natexlab{c}}.

\bibitem[Park et~al.(2023)Park, Son, Jang, Ahn, Kim, and
  Kang]{park2023temporal}
Sungheon Park, Minjung Son, Seokhwan Jang, Young~Chun Ahn, Ji-Yeon Kim, and
  Nahyup Kang.
\newblock Temporal interpolation is all you need for dynamic neural radiance
  fields.
\newblock In \emph{Proceedings of the IEEE/CVF Conference on Computer Vision
  and Pattern Recognition}, pages 4212--4221, 2023.

\bibitem[Pumarola et~al.(2021)Pumarola, Corona, Pons-Moll, and
  Moreno-Noguer]{pumarola2021d}
Albert Pumarola, Enric Corona, Gerard Pons-Moll, and Francesc Moreno-Noguer.
\newblock D-nerf: Neural radiance fields for dynamic scenes.
\newblock In \emph{Proceedings of the IEEE/CVF Conference on Computer Vision
  and Pattern Recognition}, pages 10318--10327, 2021.

\bibitem[Reiser et~al.(2023)Reiser, Szeliski, Verbin, Srinivasan, Mildenhall,
  Geiger, Barron, and Hedman]{reiser2023merf}
Christian Reiser, Rick Szeliski, Dor Verbin, Pratul Srinivasan, Ben Mildenhall,
  Andreas Geiger, Jon Barron, and Peter Hedman.
\newblock Merf: Memory-efficient radiance fields for real-time view synthesis
  in unbounded scenes.
\newblock \emph{ACM Transactions on Graphics (TOG)}, 42\penalty0 (4):\penalty0
  1--12, 2023.

\bibitem[{Sara Fridovich-Keil and Giacomo Meanti} et~al.(2023){Sara
  Fridovich-Keil and Giacomo Meanti}, Warburg, Recht, and
  Kanazawa]{kplanes_2023}
{Sara Fridovich-Keil and Giacomo Meanti}, Frederik~Rahbæk Warburg, Benjamin
  Recht, and Angjoo Kanazawa.
\newblock K-planes: Explicit radiance fields in space, time, and appearance.
\newblock In \emph{CVPR}, 2023.

\bibitem[Sch\"{o}nberger and Frahm(2016)]{schoenberger2016sfm}
Johannes~Lutz Sch\"{o}nberger and Jan-Michael Frahm.
\newblock Structure-from-motion revisited.
\newblock In \emph{Conference on Computer Vision and Pattern Recognition
  (CVPR)}, 2016.

\bibitem[Shum and He(1999)]{shum1999rendering}
Heung-Yeung Shum and Li-Wei He.
\newblock Rendering with concentric mosaics.
\newblock In \emph{Proceedings of the 26th annual conference on Computer
  graphics and interactive techniques}, pages 299--306, 1999.

\bibitem[Sloan(2008)]{sloan2008stupid}
Peter-Pike Sloan.
\newblock Stupid spherical harmonics (sh) tricks.
\newblock In \emph{Game developers conference}, page~42, 2008.

\bibitem[Song et~al.(2023)Song, Chen, Li, Chen, Chen, Yuan, Xu, and
  Geiger]{song2022nerfplayer}
Liangchen Song, Anpei Chen, Zhong Li, Zhang Chen, Lele Chen, Junsong Yuan, Yi
  Xu, and Andreas Geiger.
\newblock Nerfplayer: A streamable dynamic scene representation with decomposed
  neural radiance fields.
\newblock \emph{IEEE Transactions on Visualization and Computer Graphics},
  29\penalty0 (5):\penalty0 2732--2742, 2023.

\bibitem[Sun et~al.(2022)Sun, Sun, and Chen]{sun2021direct}
Cheng Sun, Min Sun, and Hwann-Tzong Chen.
\newblock Direct voxel grid optimization: Super-fast convergence for radiance
  fields reconstruction.
\newblock In \emph{2022 IEEE/CVF Conference on Computer Vision and Pattern
  Recognition (CVPR)}, pages 5449--5459, 2022.

\bibitem[Sun et~al.(2023)Sun, Zhang, Chen, Li, Ji, Zhao, and Xing]{sun2023vgos}
Jiakai Sun, Zhanjie Zhang, Jiafu Chen, Guangyuan Li, Boyan Ji, Lei Zhao, and
  Wei Xing.
\newblock Vgos: Voxel grid optimization for view synthesis from sparse inputs.
\newblock In \emph{Proceedings of the Thirty-Second International Joint
  Conference on Artificial Intelligence, {IJCAI-23}}, pages 1414--1422.
  International Joint Conferences on Artificial Intelligence Organization,
  2023.
\newblock Main Track.

\bibitem[Tomasi and Kanade(1992)]{tomasi1992shape}
Carlo Tomasi and Takeo Kanade.
\newblock Shape and motion from image streams under orthography: a
  factorization method.
\newblock \emph{International journal of computer vision}, 9:\penalty0
  137--154, 1992.

\bibitem[Tretschk et~al.(2021)Tretschk, Tewari, Golyanik, Zollh\"ofer, Lassner,
  and Theobalt]{tretschk2020non}
Edgar Tretschk, Ayush Tewari, Vladislav Golyanik, Michael Zollh\"ofer,
  Christoph Lassner, and Christian Theobalt.
\newblock Non-rigid neural radiance fields: Reconstruction and novel view
  synthesis of a dynamic scene from monocular video.
\newblock In \emph{Proceedings of the IEEE/CVF International Conference on
  Computer Vision (ICCV)}, pages 12959--12970, 2021.

\bibitem[Verbin et~al.(2022)Verbin, Hedman, Mildenhall, Zickler, Barron, and
  Srinivasan]{verbin2022refnerf}
Dor Verbin, Peter Hedman, Ben Mildenhall, Todd Zickler, Jonathan~T. Barron, and
  Pratul~P. Srinivasan.
\newblock {Ref-NeRF}: Structured view-dependent appearance for neural radiance
  fields.
\newblock \emph{CVPR}, 2022.

\bibitem[Wang et~al.(2023{\natexlab{a}})Wang, Tan, Li, Tian, Song, and
  Liu]{wang2023mixed}
Feng Wang, Sinan Tan, Xinghang Li, Zeyue Tian, Yafei Song, and Huaping Liu.
\newblock Mixed neural voxels for fast multi-view video synthesis.
\newblock In \emph{Proceedings of the IEEE/CVF International Conference on
  Computer Vision}, pages 19706--19716, 2023{\natexlab{a}}.

\bibitem[Wang et~al.(2022)Wang, Zhang, Liu, Zhao, Zhang, Zhang, Wu, Yu, and
  Xu]{wang2022fourier}
Liao Wang, Jiakai Zhang, Xinhang Liu, Fuqiang Zhao, Yanshun Zhang, Yingliang
  Zhang, Minye Wu, Jingyi Yu, and Lan Xu.
\newblock Fourier plenoctrees for dynamic radiance field rendering in
  real-time.
\newblock In \emph{Proceedings of the IEEE/CVF Conference on Computer Vision
  and Pattern Recognition}, pages 13524--13534, 2022.

\bibitem[Wang et~al.(2023{\natexlab{b}})Wang, Hu, He, Wang, Yu, Tuytelaars, Xu,
  and Wu]{Wang_2023_CVPR}
Liao Wang, Qiang Hu, Qihan He, Ziyu Wang, Jingyi Yu, Tinne Tuytelaars, Lan Xu,
  and Minye Wu.
\newblock Neural residual radiance fields for streamably free-viewpoint videos.
\newblock In \emph{Proceedings of the IEEE/CVF Conference on Computer Vision
  and Pattern Recognition (CVPR)}, pages 76--87, 2023{\natexlab{b}}.

\bibitem[Wang et~al.(2023{\natexlab{c}})Wang, Hu, He, Wang, Yu, Tuytelaars, Xu,
  and Wu]{wang2023neuralresidual}
Liao Wang, Qiang Hu, Qihan He, Ziyu Wang, Jingyi Yu, Tinne Tuytelaars, Lan Xu,
  and Minye Wu.
\newblock Neural residual radiance fields for streamably free-viewpoint videos.
\newblock In \emph{Proceedings of the IEEE/CVF Conference on Computer Vision
  and Pattern Recognition (CVPR)}, pages 76--87, 2023{\natexlab{c}}.

\bibitem[Wang et~al.(2023{\natexlab{d}})Wang, Chang, Cai, Li, Hariharan,
  Holynski, and Snavely]{wang2023omnimotion}
Qianqian Wang, Yen-Yu Chang, Ruojin Cai, Zhengqi Li, Bharath Hariharan,
  Aleksander Holynski, and Noah Snavely.
\newblock Tracking everything everywhere all at once.
\newblock In \emph{International Conference on Computer Vision},
  2023{\natexlab{d}}.

\bibitem[Weng et~al.(2022)Weng, Curless, Srinivasan, Barron, and
  Kemelmacher-Shlizerman]{weng2022humannerf}
Chung-Yi Weng, Brian Curless, Pratul~P Srinivasan, Jonathan~T Barron, and Ira
  Kemelmacher-Shlizerman.
\newblock Humannerf: Free-viewpoint rendering of moving people from monocular
  video.
\newblock In \emph{Proceedings of the IEEE/CVF Conference on Computer Vision
  and Pattern Recognition}, pages 16210--16220, 2022.

\bibitem[Wimbauer et~al.(2023)Wimbauer, Yang, Rupprecht, and
  Cremers]{wimbauer2023behind}
Felix Wimbauer, Nan Yang, Christian Rupprecht, and Daniel Cremers.
\newblock Behind the scenes: Density fields for single view reconstruction.
\newblock In \emph{Proceedings of the IEEE/CVF Conference on Computer Vision
  and Pattern Recognition}, pages 9076--9086, 2023.

\bibitem[Wizadwongsa et~al.(2021)Wizadwongsa, Phongthawee, Yenphraphai, and
  Suwajanakorn]{wizadwongsa2021nex}
Suttisak Wizadwongsa, Pakkapon Phongthawee, Jiraphon Yenphraphai, and Supasorn
  Suwajanakorn.
\newblock Nex: Real-time view synthesis with neural basis expansion.
\newblock In \emph{Proceedings of the IEEE/CVF Conference on Computer Vision
  and Pattern Recognition}, pages 8534--8543, 2021.

\bibitem[Wu et~al.(2023)Wu, Yi, Fang, Xie, Zhang, Wei, Liu, Tian, and
  Xinggang]{wu20234dgaussians}
Guanjun Wu, Taoran Yi, Jiemin Fang, Lingxi Xie, Xiaopeng Zhang, Wei Wei, Wenyu
  Liu, Qi Tian, and Wang Xinggang.
\newblock 4d gaussian splatting for real-time dynamic scene rendering.
\newblock \emph{arXiv preprint arXiv:2310.08528}, 2023.

\bibitem[Wynn and Turmukhambetov(2023)]{wynn2023diffusionerf}
Jamie Wynn and Daniyar Turmukhambetov.
\newblock Diffusionerf: Regularizing neural radiance fields with denoising
  diffusion models.
\newblock In \emph{Proceedings of the IEEE/CVF Conference on Computer Vision
  and Pattern Recognition}, pages 4180--4189, 2023.

\bibitem[Xian et~al.(2021)Xian, Huang, Kopf, and Kim]{xian2020space}
Wenqi Xian, Jia-Bin Huang, Johannes Kopf, and Changil Kim.
\newblock Space-time neural irradiance fields for free-viewpoint video.
\newblock In \emph{Proceedings of the IEEE/CVF Conference on Computer Vision
  and Pattern Recognition (CVPR)}, pages 9421--9431, 2021.

\bibitem[Yang et~al.(2022)Yang, Vo, Neverova, Ramanan, Vedaldi, and
  Joo]{yang2022banmo}
Gengshan Yang, Minh Vo, Natalia Neverova, Deva Ramanan, Andrea Vedaldi, and
  Hanbyul Joo.
\newblock Banmo: Building animatable 3d neural models from many casual videos.
\newblock In \emph{Proceedings of the IEEE/CVF Conference on Computer Vision
  and Pattern Recognition}, pages 2863--2873, 2022.

\bibitem[Yang et~al.(2023{\natexlab{a}})Yang, Pavone, and
  Wang]{yang2023freenerf}
Jiawei Yang, Marco Pavone, and Yue Wang.
\newblock Freenerf: Improving few-shot neural rendering with free frequency
  regularization.
\newblock In \emph{Proceedings of the IEEE/CVF Conference on Computer Vision
  and Pattern Recognition}, pages 8254--8263, 2023{\natexlab{a}}.

\bibitem[Yang et~al.(2023{\natexlab{b}})Yang, Gao, Zhou, Jiao, Zhang, and
  Jin]{yang2023deformable3dgs}
Ziyi Yang, Xinyu Gao, Wen Zhou, Shaohui Jiao, Yuqing Zhang, and Xiaogang Jin.
\newblock Deformable 3d gaussians for high-fidelity monocular dynamic scene
  reconstruction.
\newblock \emph{arXiv preprint arXiv:2309.13101}, 2023{\natexlab{b}}.

\bibitem[Yang et~al.(2024)Yang, Yang, Pan, and Zhang]{yang2023gs4d}
Zeyu Yang, Hongye Yang, Zijie Pan, and Li Zhang.
\newblock Real-time photorealistic dynamic scene representation and rendering
  with 4d gaussian splatting.
\newblock 2024.

\bibitem[Yu et~al.(2021{\natexlab{a}})Yu, Li, Tancik, Li, Ng, and
  Kanazawa]{yu2021plenoctrees}
Alex Yu, Ruilong Li, Matthew Tancik, Hao Li, Ren Ng, and Angjoo Kanazawa.
\newblock Plenoctrees for real-time rendering of neural radiance fields.
\newblock In \emph{Proceedings of the IEEE/CVF International Conference on
  Computer Vision}, pages 5752--5761, 2021{\natexlab{a}}.

\bibitem[Yu et~al.(2021{\natexlab{b}})Yu, Ye, Tancik, and
  Kanazawa]{yu2020pixelnerf}
Alex Yu, Vickie Ye, Matthew Tancik, and Angjoo Kanazawa.
\newblock {pixelNeRF}: Neural radiance fields from one or few images.
\newblock In \emph{CVPR}, 2021{\natexlab{b}}.

\bibitem[Zhao et~al.(2022)Zhao, Yang, Zhang, Lin, Zhang, Yu, and
  Xu]{zhao2022humannerf}
Fuqiang Zhao, Wei Yang, Jiakai Zhang, Pei Lin, Yingliang Zhang, Jingyi Yu, and
  Lan Xu.
\newblock Humannerf: Efficiently generated human radiance field from sparse
  inputs.
\newblock In \emph{Proceedings of the IEEE/CVF Conference on Computer Vision
  and Pattern Recognition}, pages 7743--7753, 2022.

\bibitem[Zitnick et~al.(2004)Zitnick, Kang, Uyttendaele, Winder, and
  Szeliski]{zitnick2004high}
C~Lawrence Zitnick, Sing~Bing Kang, Matthew Uyttendaele, Simon Winder, and
  Richard Szeliski.
\newblock High-quality video view interpolation using a layered representation.
\newblock \emph{ACM transactions on graphics (TOG)}, 23\penalty0 (3):\penalty0
  600--608, 2004.

\bibitem[Zwicker et~al.(2001)Zwicker, Pfister, Van~Baar, and
  Gross]{zwicker2001ewa}
Matthias Zwicker, Hanspeter Pfister, Jeroen Van~Baar, and Markus Gross.
\newblock Ewa volume splatting.
\newblock In \emph{Proceedings Visualization, 2001. VIS'01.}, pages 29--538.
  IEEE, 2001.

\end{thebibliography}
